%% file: neurips_2025.tex
\documentclass{article}

\PassOptionsToPackage{numbers, compress}{natbib}

\usepackage[final]{neurips_2025}

\usepackage[utf8]{inputenc} 
\usepackage[T1]{fontenc}    
\usepackage{url}            
\usepackage{booktabs}       

\usepackage{amsfonts}       
\usepackage{nicefrac}       
\usepackage{microtype}      
\usepackage{xcolor}         
\usepackage{graphicx}
\usepackage{enumitem}
\usepackage{amsmath}
\usepackage{wrapfig}        
\usepackage{gradient-text}
\usepackage{setspace}
\usepackage{multirow}
\usepackage{algorithm}    
\usepackage{algpseudocode}
\usepackage{colortbl}
\usepackage{soul}
\usepackage{caption}
\usepackage{makecell}
\usepackage[normalem]{ulem}

\definecolor{blue}{rgb}{0.21,0.49,0.74}
\usepackage[colorlinks=true, linkcolor=blue, citecolor=blue, urlcolor=blue]{hyperref}
\usepackage[capitalise]{cleveref}       

\title{%
  \raisebox{-.18\height}{\includegraphics[width=0.07\textwidth]{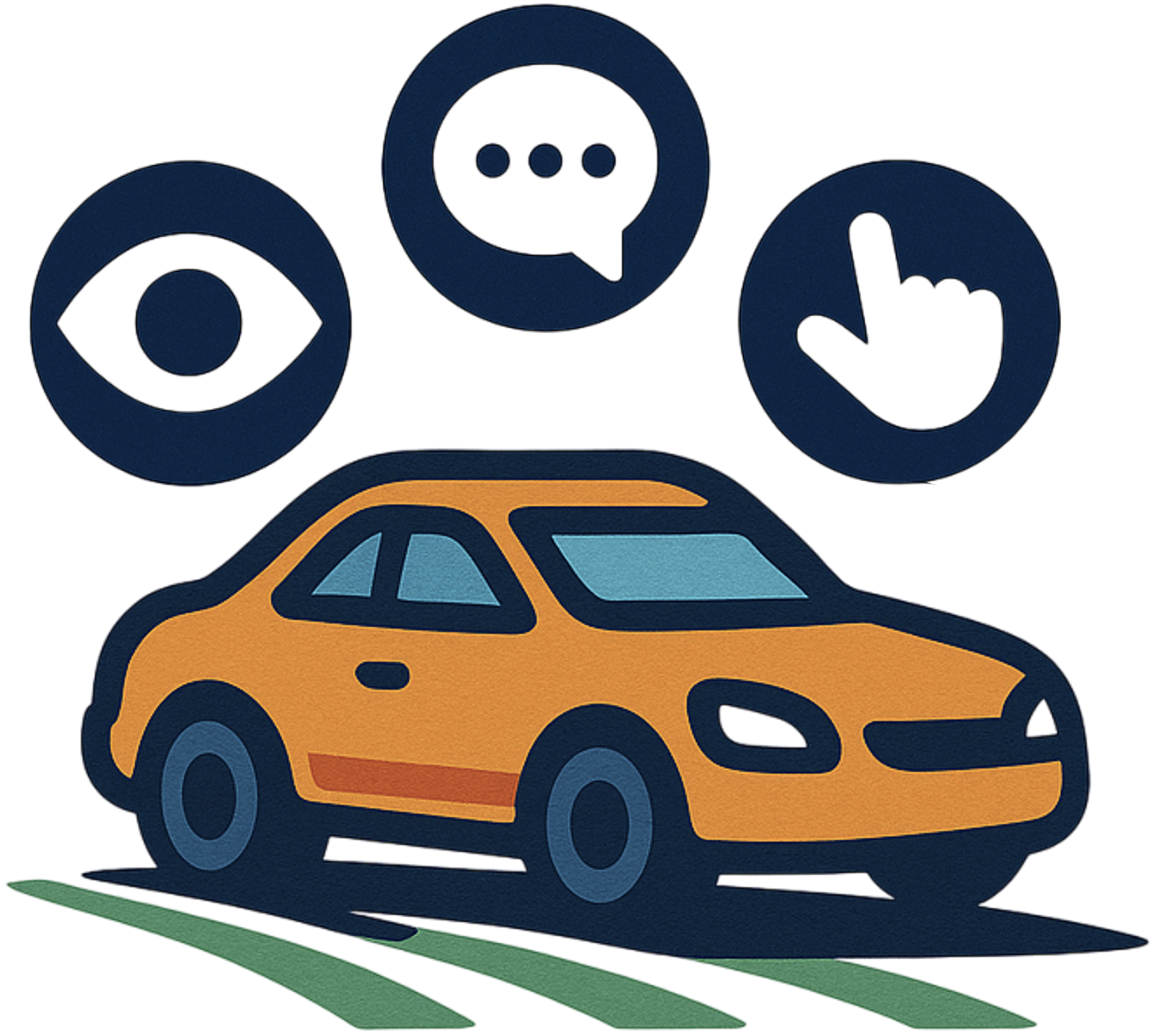}} \hspace{0.15em}%
  \textit{AutoVLA}: A Vision-Language-Action Model for End-to-End Autonomous Driving with Adaptive Reasoning and Reinforcement Fine-Tuning
}
\author{
Zewei Zhou\thanks{Equal contribution. Email: \texttt{\{zeweizhou, tianhui\}@ucla.edu}}\quad 
Tianhui Cai\footnotemark[1]\quad  
Seth Z. Zhao\:
Yun Zhang\:
Zhiyu Huang\thanks{Corresponding author. Email: \texttt{zhiyuh@ucla.edu}}\quad
Bolei Zhou\:
Jiaqi Ma \\ [0.1cm] 
University of California, Los Angeles
\\ [0.1cm]
\small \tt{\href{https://autovla.github.io/}{https://autovla.github.io/}}
}

\begin{document}

\maketitle

\vspace{-0.3cm}
\begin{abstract}

Recent advancements in Vision-Language-Action (VLA) models have shown promise for end-to-end autonomous driving by leveraging world knowledge and reasoning capabilities. However, current VLA models often struggle with physically infeasible action outputs, complex model structures, or unnecessarily long reasoning. In this paper, we propose \textbf{AutoVLA}, a novel VLA model that unifies reasoning and action generation within a single autoregressive generation model for end-to-end autonomous driving. AutoVLA performs semantic reasoning and trajectory planning directly from raw visual inputs and language instructions. We tokenize continuous trajectories into discrete, feasible actions, enabling direct integration into the language model. For training, we employ supervised fine-tuning to equip the model with dual thinking modes: fast thinking (trajectory-only) and slow thinking (enhanced with chain-of-thought reasoning). To further enhance planning performance and efficiency, we introduce a reinforcement fine-tuning method based on Group Relative Policy Optimization (GRPO), reducing unnecessary reasoning in straightforward scenarios. Extensive experiments across real-world and simulated datasets and benchmarks, including nuPlan, nuScenes, Waymo, and CARLA, demonstrate the competitive performance of AutoVLA in both open-loop and closed-loop settings. Qualitative results showcase the adaptive reasoning and accurate planning capabilities of AutoVLA in diverse scenarios. 

\end{abstract}

\section{Introduction}
\label{intro}

Autonomous driving systems typically adopt a modular paradigm, decomposing the driving task into different sub-modules, such as perception \cite{li2024bevformer, wang2022detr3d, liang2022bevfusion}, prediction \cite{zhou2023qcnext, shi2024mtr++, huang2023gameformer}, and planning \cite{huang2024gen, huang2023differentiable, liu2025hybrid}. While this design enables structured development, it may cause error accumulation and a lack of joint optimization across modules, leading to suboptimal performance \cite{shao2023reasonnet, jia2023think}. End-to-end autonomous driving has gained prominence with a unified model architecture that maps raw sensor inputs directly to final driving actions. These models are trained on human driving data, enhancing scalability and human-like behavior. Vision-based approaches have garnered significant interest due to their affordability and ease of deployment \cite{pan2024vlp, Wang_2024_CVPR, hu2022st, fang2023tbp}. 

However, conventional end-to-end methods \cite{zhou2024vision, cai2024driving, chen2024end, fu2024drive} primarily focus on imitating expert trajectories, lacking essential world knowledge for understanding and reasoning about surrounding environments, particularly in long-tail or challenging scenarios. Recent advances in Vision-Language Models (VLMs) \cite{jaech2024openai, bai2025qwen2, guo2025deepseek} have gained significant interest by introducing models capable of leveraging extensive world knowledge and powerful reasoning. These models have shown strong potential in improving adaptability and scalability across diverse driving scenarios \cite{chen2025automated, li2024driving, xu2024drivegpt4, wang2024omnidrive, ma2024dolphins, winter2025bevdriver, tian2025nuscenes}. Building upon VLMs, Vision-Language-Action (VLA) models extend this capability to action generation, enabling embodied agents, such as robots \cite{kim2024openvla, zhao2025cot, hung2025norasmallopensourcedgeneralist} and autonomous vehicles \cite{arai2025covla, zhou2025opendrivevla}, to produce feasible physical actions based on visual observations and language instructions.

\begin{figure}[t]
    \centering
    \includegraphics[width=0.99\linewidth]{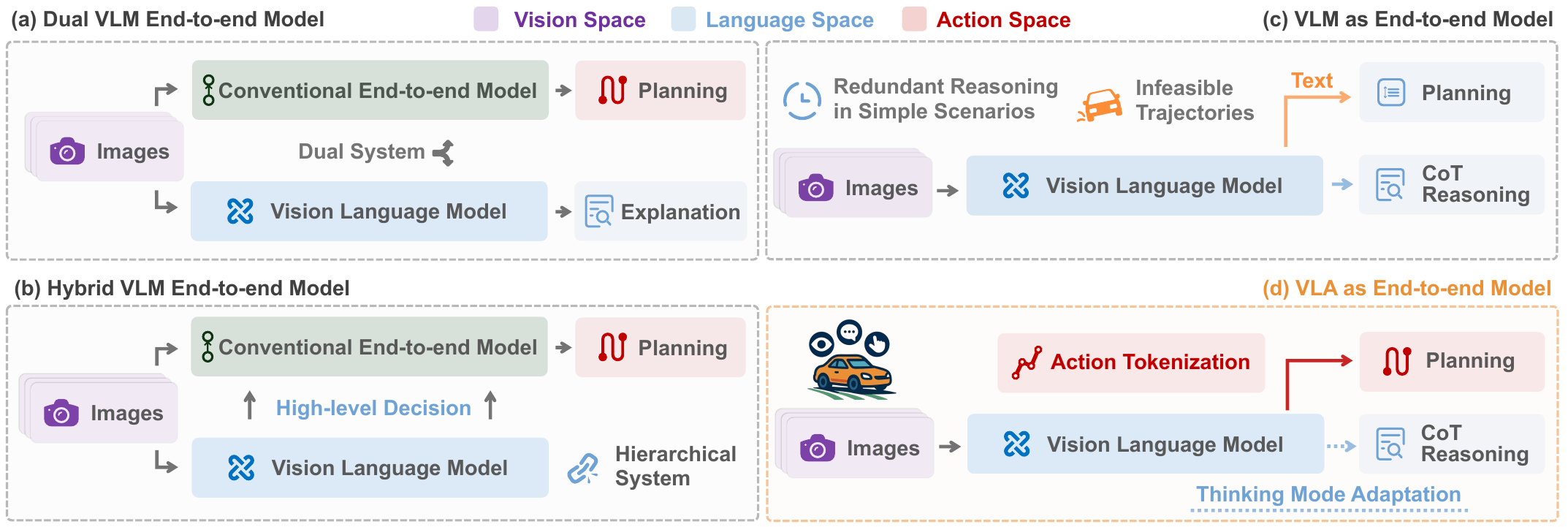}    
    \caption{Four paradigms of VLMs for end-to-end autonomous driving. Compared to other methods, our proposed VLA-based paradigm enables direct trajectory planning and adaptive reasoning from visual inputs. By incorporating physical action tokenization into the language model, our model effectively integrates high-level scene reasoning and low-level trajectory planning.}
    \label{fig:2}
    \vspace{-0.4cm}
\end{figure}

Despite recent progress, existing VLA models face two critical limitations in autonomous driving, as illustrated in \cref{fig:2}. \textit{1) Physically-infeasible or complex structure for action generation.} Some models generate textual actions or waypoints directly using VLMs \cite{shao2024lmdrive, renz2025simlingo, hwang2024emma}, but these outputs can be physically infeasible and suffer from mode collapse. To address this, recent approaches introduce intermediate meta-actions \cite{jiang2025alphadrive, jiang2024senna, wang2023drivemlm} or latent action tokens \cite{renz2024carllava, fu2025orion, waywe2024lingo}, which are then processed by downstream planners or decoders to produce physically feasible trajectories. However, the intermediate representations either break the end-to-end optimization paradigm or increase model complexity and training overhead. \textit{2) Inflexible and inefficient reasoning across diverse scenarios.} Most existing models \cite{xing2025openemma, wang2024drivecot} employ a fixed reasoning strategy, lacking the ability to adaptively switch between direct action outputs for straightforward scenarios and chain-of-thought (CoT) reasoning for complex ones. Although DriveVLM \cite{tian2024drivevlm} introduces a dual-process paradigm, it relies on separate modules (i.e., a VLM for slow reasoning and a conventional end-to-end model for fast responses), which results in a complicated architecture, increased training overhead, and limited scalability \cite{ma2025reasoning}.

\begin{figure}[t]
    \centering
    \includegraphics[width=0.99\linewidth]{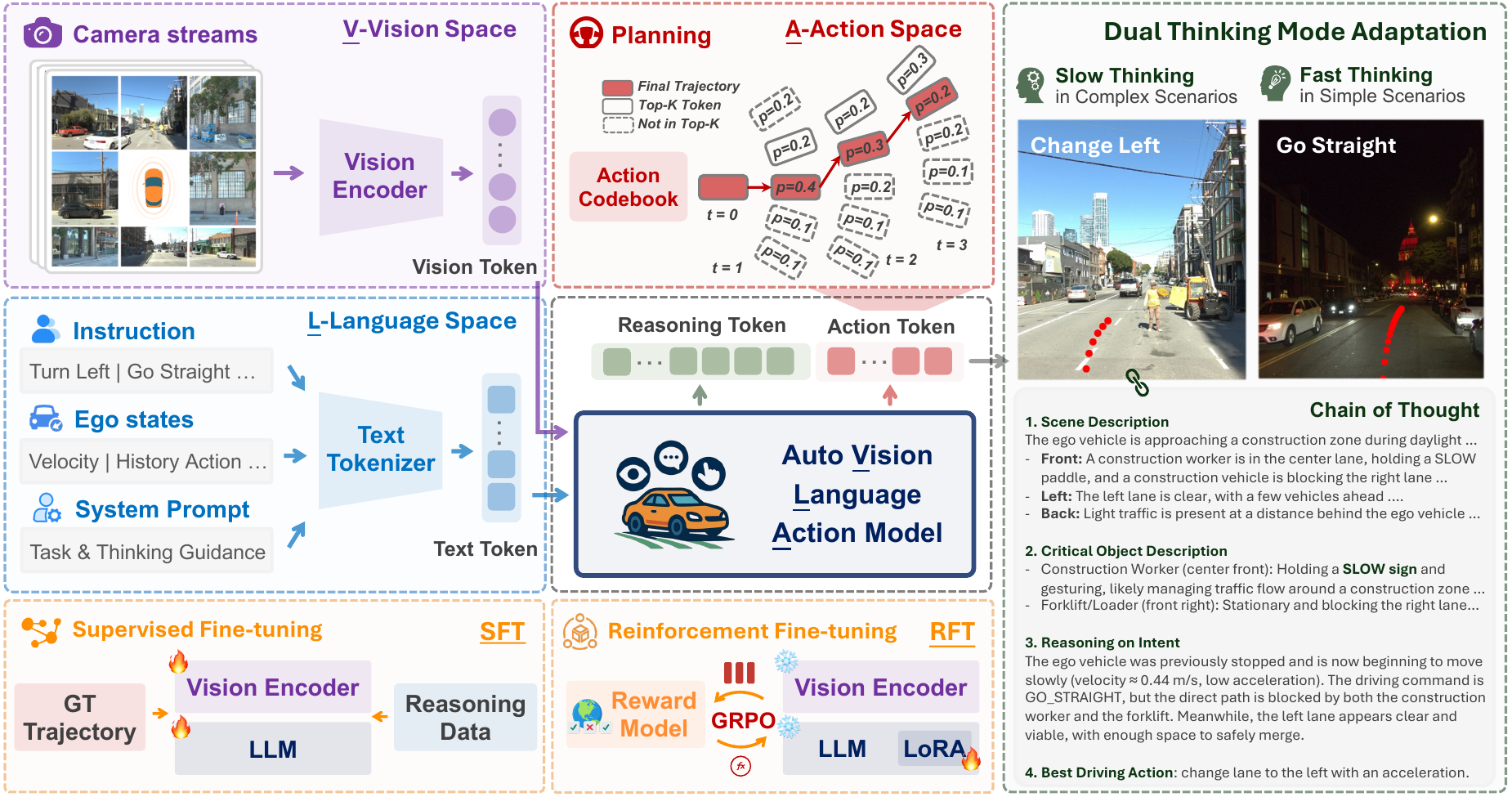}
    \caption{AutoVLA is an end-to-end autonomous driving framework based on vision-language models that integrates world knowledge into the driving policy. It takes visual observations, vehicle states, and language instructions as input and incorporates CoT reasoning and physical action tokenization to generate planning trajectories directly. The model is trained using supervised fine-tuning to jointly learn reasoning and action, and reinforcement fine-tuning is further applied to enable adaptive reasoning through fast and slow thinking modes, improving performance and efficiency.}
    \label{fig:1}
    \vspace{-0.3cm}
\end{figure}

To overcome these limitations, we propose \textbf{AutoVLA}, an end-to-end autonomous driving framework that directly integrates physical action tokens into a pretrained VLM backbone, enabling direct learning of an autoregressive planning policy, as illustrated in \cref{fig:1}. Our unified architecture seamlessly integrates reasoning and action generation, allowing adaptive switching between direct trajectory generation and CoT reasoning. In supervised fine-tuning (SFT), we leverage both trajectory-only data and CoT reasoning data to equip the model with dual-process capabilities (fast and slow thinking). Furthermore, we propose reinforcement fine-tuning (RFT) \cite{ouyang2022training}, utilizing Group Relative Policy Optimization (GRPO) \cite{shao2024deepseekmath} with verifiable planning reward functions. This enables adaptive reasoning that balances planning accuracy and efficiency. The RFT method not only improves planning performance but also runtime efficiency by minimizing unnecessary reasoning.

We extensively evaluate AutoVLA using real-world datasets, including nuPlan \cite{karnchanachari2024towards, dauner2024navsim}, Waymo \cite{xu2025wod}, nuScenes \cite{caesar2020nuscenes}, and simulation datasets such as CARLA \cite{jia2024bench2drive, Jaeger_2023_ICCV}. Experimental results demonstrate that AutoVLA achieves superior performance across various end-to-end autonomous driving benchmarks under both open-loop and closed-loop tests. Empirical results validate that our RFT approach markedly improves planning performance, enables adaptive fast and slow thinking capabilities, and reduces runtime by minimizing redundant reasoning. The main contributions of this paper are summarized as follows:
\begin{enumerate}[leftmargin=1.5em] 
\item We introduce AutoVLA, an end-to-end autonomous driving framework leveraging a pretrained VLM backbone integrated with physical action tokens, enabling direct policy learning and semantic reasoning from raw visual observations and language instructions.
\item We propose an RL-based post-training method using GRPO to enable adaptive reasoning and further enhance the model's performance on end-to-end driving tasks. 
\item We demonstrate that AutoVLA achieves superior performance across multiple autonomous driving benchmarks, including both open-loop and closed-loop testing. 
\end{enumerate}

\section{Related Work}
\label{related}

\noindent \textbf{End-to-end Autonomous Driving.} 
End-to-end autonomous driving approaches have made significant advances in recent years \cite{shao2023reasonnet, jia2023think, yuan2024drama,  li2024hydra, chitta2022transfuser, sima2025centaur, zhou2024v2xpnp, sun2024sparsedrive, li2024pretrain, zheng2024gaussianad, wang2025generativeaiautonomousdriving}. Methods such as UniAD \cite{hu2023planning} and VAD \cite{jiang2023vad} explicitly integrate multiple driving tasks from perception to planning in a unified Transformer architecture, thereby enhancing planning performance. ParaDrive \cite{weng2024drive} discusses the necessary components within end-to-end driving architectures. Additionally, GenAD \cite{zheng2024genad} and DiffusionDrive \cite{liao2024diffusiondrive} adopt generative models to maintain trajectory continuity and produce multi-modal driving trajectories.
However, integrating world knowledge into end-to-end driving systems remains challenging due to bottlenecks in semantic reasoning \cite{zhou2025opendrivevla} and limited adaptability in complex environments \cite{li2024ego}.

\textbf{VLA and VLM for Autonomous Driving.} 
The gap between semantic reasoning and physical actions remains a critical challenge for VLA and VLM in end-to-end autonomous driving. Current research broadly follows three directions. The first directly formulates driving as a language-centric problem, utilizing VLMs for scenario understanding through caption generation \cite{jin2023adapt, park2024vlaad, ding2024hint} or question answering \cite{marcu2024lingoqa, li2024womd}. The second direction leverages VLA or VLM to produce high-level meta-actions or driving decisions \cite{cai2024driving, jiang2025alphadrive, jiang2024senna, wang2023drivemlm}, which are used to either supervise \cite{pan2024vlp,xu2024vlm,hegde2025distilling,liu2025vlm} or guide \cite{jiang2024senna, liao2025cot} traditional planners or end-to-end models. Although these approaches facilitate integration, they prevent full end-to-end optimization. Thus, a third direction directly integrates VLMs with action generation into VLA models, enabling the direct prediction of latent action tokens \cite{zhou2025opendrivevla, shao2024lmdrive, renz2025simlingo, waywe2024lingo} or final driving trajectories \cite{hwang2024emma, xing2025openemma, mao2023gpt, zhang2024wisead,tian2024tokenize, qiao2025lightemmalightweightendtoendmultimodal}. However, simple trajectory decoders employed in these methods (e.g., MLP \cite{renz2024carllava, liu2025dsdrivedistillinglargelanguage} or GRU \cite{wang2024drivecot}) may produce impractical trajectories and suffer from modal collapse. To address this issue, ORION \cite{fu2025orion} incorporates generative planners into VLM architectures, enhancing trajectory feasibility but increasing model complexity and computational demands. In our work, we integrate a physical action codebook for vehicle motion into a pretrained VLM to effectively bridge the semantic reasoning and physical action space.

\textbf{Reinforcement Fine-tuning.} 
RFT \cite{ouyang2022training} has shown considerable promise in enhancing the performance and adaptability of LLMs, as demonstrated in DeepSeek-R1 \cite{guo2025deepseek}. In autonomous driving, Gen-Drive \cite{huang2024gen} and TrajHF \cite{li2025finetuning} employed the RFT to align the trajectory generation model with safety constraints and human driving preferences. RAD \cite{gao2025rad} combined 3D Gaussian splatting to generate scenarios and conduct closed-loop RL training. However, the application of RFT in end-to-end VLM/VLA-based autonomous driving remains nascent. While previous methods, such as AlphaDrive \cite{jiang2025alphadrive}, utilize GRPO instead of direct preference optimization (DPO) \cite{rafailov2023direct} to enhance planning performance and ensure training efficiency and stability, they are still limited to simplified settings involving only high-level meta-actions. In this work, we advance this direction by applying RFT to optimize the end-to-end VLA framework in both scene reasoning and low-level planning, and we adopt GRPO to accelerate convergence and enhance training stability.

\begin{figure}[t]
    \centering
    \includegraphics[width=0.99\linewidth]{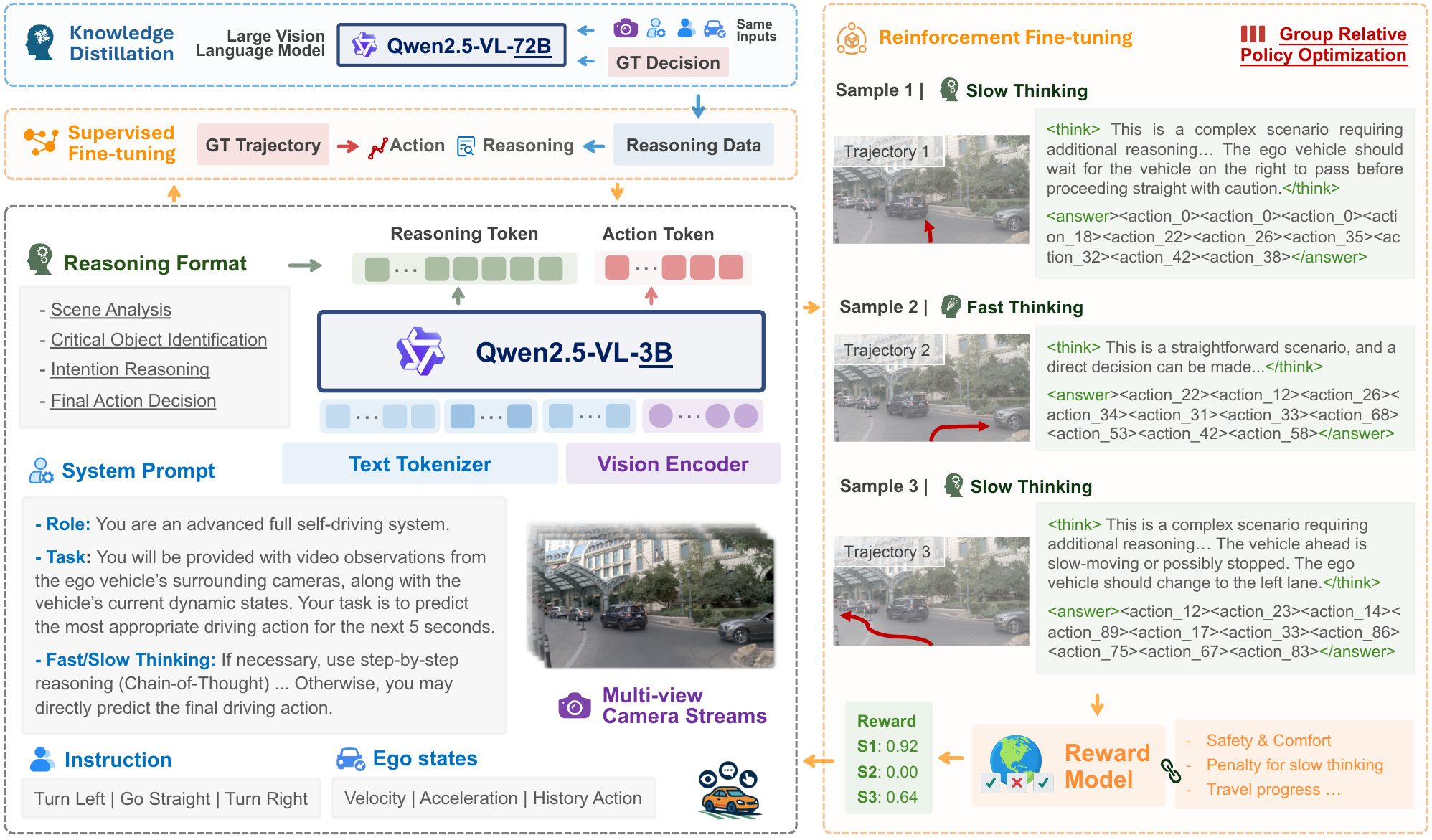}
    \caption{Overview of the AutoVLA model and its training process. A pretrained small VLM is used as the backbone of AutoVLA. The model receives multi-view camera streams, system prompts, driving instructions, and vehicle status as input, and outputs textual reasoning and physical action tokens. In SFT, a large VLM model with strong visual understanding capabilities is employed to collect reasoning data, which is used alongside trajectory data in SFT for training the AutoVLA model. In RFT, we utilize GRPO to train the model for improved alignment with verified reward functions, while enabling adaptive reasoning by penalizing excessive reasoning.}
    \label{fig:3}
    \vspace{-0.4cm}
\end{figure}

\section{AutoVLA}
\label{method}

The proposed AutoVLA framework consists of two main components, as shown in \cref{fig:1}.
\textit{1) VLM Backbone:} It is capable of processing visual and textual input and generating corresponding tokens (reasoning and action), employing a unified autoregressive Transformer decoder.
\textit{2) Physical Action Token Generation:} We extend the language model decoder to output physical action tokens that directly correspond to vehicle movements. These tokens are designed to comply with physical constraints and can be reliably translated into physically feasible planning trajectories.

Training of AutoVLA is conducted in two stages, as illustrated in \cref{fig:3}.
\textit{1) Supervised Fine-Tuning} uses ground-truth trajectory data and distills high-quality reasoning data from a large-scale VLM.
\textit{2) Reinforcement Fine-Tuning} uses task-specific reward functions to optimize planning performance while improving the running efficiency by minimizing unnecessary reasoning. The details of our model and training process are illustrated below.

\subsection{Framework}

\noindent \textbf{Model Inputs.}
AutoVLA takes as input multi-view, multi-frame camera data $C$ from onboard cameras, high-level navigation instructions $I$, and ego vehicle states $S$, and performs scene reasoning and trajectory planning. Specifically, we utilize three RGB cameras positioned at the front, front-left, and front-right sides of the vehicle. Each camera stream $c^i = [c^i_{t-3}, c^i_{t-2}, c^i_{t-1}, c^i_t]$ captures four sequential frames at a frequency of 2 Hz, including the current and three preceding frames, supplying temporal information for scene dynamics. Additionally, the model employs high-level navigation instructions $I$ (e.g., Turn Left and Go Straight) to specify intended directions explicitly. The ego vehicle's state $S$ encompasses current velocity, acceleration, and historical actions. 

\noindent \textbf{Base VLM Model.}
We adopt Qwen2.5-VL-3B \cite{bai2025qwen2} as the vision-language backbone of AutoVLA. Qwen2.5-VL is a series of powerful multimodal large language models that possess strong visual understanding capabilities, and the open-source nature of the Qwen2.5-VL model facilitates task-specific fine-tuning. The 3B variant offers a good trade-off between efficiency and performance, making it suitable for deployment in onboard devices.

\noindent \textbf{Action Tokenization.}
To enable trajectory planning within the language model, we discretize continuous vehicle trajectories $\mathbf{P} \in \mathbb{R}^{\tau \times d}$ into a sequence of physical action tokens $\mathbf{a} = [a_1, \dots, a_T]$, where $a_t \in \mathcal{A}$, $T$ is the length of the tokenized predicted trajectory and each token is represented by short-term spatial position and heading movement $(\Delta x, \Delta y, \Delta \theta)$. This transforms the planning task into a next-token prediction problem, which can be conducted within the language model. We build our action codebook $\mathcal{A} = \{a_1, a_2, \dots, a_K\}$ using a K-disk clustering method \cite{zhang2024closed, philion2023trajeglish, wu2024smart}, which covers the majority of vehicle movement patterns. Finally, we obtain a vehicle motion codebook that consists of $K = 2048$ discrete action tokens. Following \cite{kim2024openvla, pertsch2025fast}, these action tokens are incorporated into the VLM as additional tokens (i.e., \texttt{<action\_0>}, \texttt{<action\_1>}, \ldots). During inference, the model outputs a sequence of these action tokens, which are subsequently decoded into a planning trajectory using the action codebook. More details about action tokenization and trajectory decoding are provided in the Supplementary Material. 

\noindent \textbf{Unified Reasoning and Action.}
AutoVLA unifies reasoning and action generation within a single autoregressive Transformer framework, enabling adaptive switching between fast and slow thinking depending on the driving scenario. In fast thinking mode, AutoVLA directly predicts physical action tokens without generating long CoT reasoning, enabling rapid responses in straightforward scenarios. In contrast, slow thinking mode involves structured CoT reasoning, where the model first analyzes the environment, identifies critical elements, and reasons through potential outcomes before deciding on the final driving action. To enable this dual thinking capability, AutoVLA is trained with a mixture of direct action supervision and reasoning-augmented data. We design system prompts and response formats to support both modes consistently. 

\subsection{Reasoning Data}

Reasoning data provides high-quality CoT annotations that are essential for training VLMs with reasoning capabilities \cite{fu2025orion}. In driving tasks, reasoning involves understanding complex semantics and interactions in dynamic environments \cite{park2025nuplanqa, wu2025language,qian2024nuscenes, nie2024reason2drive}. Despite its importance, the development of a high-quality, large-scale driving reasoning dataset remains a key challenge due to three major limitations: 1) limited scenario diversity and repetitive examples, 2) inadequate representation of critical perceptual cues, such as traffic signs and vehicle indicator signals, 3) low-quality reasoning process, such as repeatedly stopping at a stop sign without justification.

To address these issues, we propose an automated reasoning annotation pipeline using the advanced Qwen2.5-VL-72B model \cite{bai2025qwen2}. This pipeline enables automatic generation of high-accuracy reasoning annotations and supports knowledge distillation from a large capable model to a more compact target model. The pipeline generates structured reasoning annotations across four key components: detailed scene descriptions, identification of crucial objects, prediction of surrounding agents' intentions, and determination of appropriate driving actions. To regulate the reasoning outcomes, our approach incorporates ground-truth driving actions as hints, guiding the model to produce causal explanations that explicitly link driving decisions to scene context. This structured prompting method significantly reduces nonsensical outputs and minimizes the need for manual correction.

Employing this annotation pipeline, we compile a comprehensive reasoning dataset comprising approximately 45.6k CoT reasoning annotations for the nuPlan dataset and 7.2k annotations for the Waymo end-to-end driving dataset. Additionally, we reformat and integrate DriveLM \cite{sima2024drivelm}, a VQA dataset built on nuScenes and CARLA simulation data, to augment our reasoning data. Additional details and illustrative examples are provided in the Supplementary Material.

\subsection{Supervised Fine-tuning}

Supervised fine-tuning (SFT) is employed to train the model to generate both reasoning and action sequences. Given multi-frame camera images $C$, a high-level navigation instruction $I$, and the ego vehicle state $S$, the model is trained to produce a sequence of output tokens. The output sequence consists of language tokens $\mathbf{l} = [l_1, \dots, l_L]$ for reasoning followed by action tokens $\mathbf{a} = [a_1, \dots, a_T]$. To enable both fast and slow thinking during SFT, we curate training data with ground-truth assistant responses that either include only the final action tokens or combine CoT reasoning with the corresponding action tokens. In the fast-thinking mode, $\mathbf{l}$ is a fixed, short template indicating that reasoning is not needed. Conversely, in the slow-thinking mode, $\mathbf{l}$ begins with a template that introduces the need for CoT reasoning, followed by a structured sequence of reasoning.

The first supervision signal is the standard causal language modeling objective, which minimizes the negative log-likelihood of the target token sequence and facilitates the reasoning capability. The other supervision signal focuses on the planning accuracy, and we introduce an auxiliary loss over action tokens $\mathbf{a} = [a_1, \dots, a_T]$, which appear at positions $x_{L+1}$ to $x_{L+T}$ in the output sequence. Given an output sequence $\mathbf{x} = [l_1, \dots, l_L, a_1, \dots, a_T]$, the loss functions are defined as:
\begin{equation}
\mathcal{L}_{\text{LM}} = - \frac{1}{N} \sum_{i=1}^{N} \log p_\theta(x_i \mid x_{<i}, C, I, S), \quad \mathcal{L}_{\text{action}} = - \frac{1}{T} \sum_{i=L+1}^{L+T} \log p_\theta(x_i \mid x_{<i}, C, I, S),
\end{equation}
where $N = L + T$, and $p_\theta$ denotes the model’s predicted distribution parameterized by $\theta$.

To jointly optimize reasoning and action generation, we combine the language modeling loss and the action loss into a single SFT loss function. To address the imbalance between reasoning data and action-only data, and to encourage the model to learn from examples that include CoT reasoning, we apply a per-sample weighting factor based on the presence of CoT in the ground truth. The overall loss for each training example is computed as follows:
\begin{equation}
 \mathcal{L}_i^\text{SFT} = w_i \cdot \left( \mathcal{L}_{\text{LM},i} + \lambda_{\text{a}} \mathcal{L}_{\text{action},i} \right), \quad
w_i =
\begin{cases}
\lambda_{\text{cot}} & \text{if CoT is present in GT} \\
1 & \text{otherwise}
\end{cases},
\end{equation}
where $\lambda_{\text{a}}$ and $\lambda_{\text{cot}}$ are hyperparameters that control the relative importance.

\subsection{Reinforcement Fine-tuning}

To further improve the performance of AutoVLA and align it with driving requirements and task-specific rewards, we introduce a reinforcement learning-based post-training method. This RFT stage enables the model to perform adaptive reasoning and optimize planning performance. We employ the GRPO algorithm \cite{shao2024deepseekmath}, which stabilizes training and improves convergence efficiency. Moreover, the inherent multi-modality of planning, characterized by multiple feasible trajectories in the same scenario, naturally aligns with the group-based optimization framework of GRPO \cite{jiang2025alphadrive}.

Given a scenario input query $q$, comprising sensor images, the ego vehicle's state, and driving instruction, we sample a set of $G$ candidate outputs $O = \{o_1, o_2, \ldots, o_G\}$ from the old policy $\pi_{\theta_{\text{old}}}$. The current policy $\pi_{\theta}$ is then optimized using the normalized group-relative advantage $A_i$, by maximizing the following objective:
\begin{equation}
\label{grpo}
\mathcal{J}_{\text{GRPO}}(\theta) = \mathbb{E}_{q, \{o_i\} \sim \pi_{\theta_{\text{old}}}(O|q)} \left[
\frac{1}{G} \sum_{i=1}^{G} \left(\mathcal{J}^{R}_i
- \beta \mathbb{D}_{\text{KL}}(\pi_\theta \| \pi_{\text{ref}})
\right)
\right],
\end{equation}
\begin{equation}
\mathcal{J}_i^{R} = 
\min \left(
\frac{\pi_\theta(o_i|q)}{\pi_{\theta_{\text{old}}}(o_i|q)} A_i,\ 
\text{clip} \left( \frac{\pi_\theta(o_i|q)}{\pi_{\theta_{\text{old}}}(o_i|q)}, 1 - \epsilon, 1 + \epsilon \right) A_i
\right), \
A_i = \frac{r_i - \text{mean}(\{r_j\}_{j=1}^G)}
{\text{std}(\{r_j\}_{j=1}^G)},
\end{equation}
where $\theta$ and $\theta_{old}$ denote the current and old policy parameters, $r_i$ is the reward for sample $o_i$, $\epsilon$ and $\beta$ are hyperparameters controlling the clipping range and the weight of the KL divergence regularization term, and $\pi_{\text{ref}}$ is the reference policy from the SFT stage. 

The final reward function is defined as $r=r_{\text{Driving}}-\lambda_{r}r_{\text{CoT}}$, where the $\lambda_{r}$ denotes the balance weight. The term $r_{\text{Driving}}$ varies across benchmarks. For the nuPlan dataset, we employ the Predictive Driver Model Score (PDMS) \cite{dauner2024navsim} as the driving reward, which captures aspects such as safety, comfort, travel efficiency, and other driving quality metrics. For the Waymo end-to-end driving dataset, due to the limited availability of Rater Feedback Score (RFS) annotations \cite{xu2025wod}, we use the Average Displacement Error (ADE) as the driving reward. To discourage unnecessary long reasoning chains, we incorporate a CoT length penalty $r_{\text{CoT}}$ into the reward function. Additional implementation details are provided in the Supplementary Material.

\begin{table}[t]
  \vspace{-0.3cm}
  \caption{Testing Results on the NAVSIM (nuPlan) End-to-end Driving Benchmark}
  \vspace{0.2cm}
  \label{navsim}
  \centering
  \small
  \setlength{\tabcolsep}{4pt}
  \begin{tabular}{l|ccccccc}
    \toprule
    Method                                      & PDMS $\uparrow$   & Collision $\uparrow$ & Area $\uparrow$   & Direction $\uparrow$ & Progress $\uparrow$ & TTC $\uparrow$    & Comfort $\uparrow$ \\
    \midrule
    Ego Status MLP                              & 66.40  & 93.09     & 78.26  & 90.45     & 63.20    & 84.02  & 99.97 \\
    TransFuser \cite{chitta2022transfuser}      & 83.88  & 97.78     & 92.63  & 97.97     & 78.88    & 92.89  & 99.98 \\
    DRAMA \cite{yuan2024drama}                  & 86.87  & 98.19     & 95.18  & \textbf{98.03} & 81.33    & 94.17  & 100.00 \\
    Hydra-MDP \cite{li2024hydra}                & 91.26  & 99.07     & 98.29  & 95.79     & 85.20    & 96.56  & \textbf{100.00} \\
    Centaur \cite{sima2025centaur}              & 92.10  & 99.23     & \textbf{98.72} & 96.77     & 85.96    & 97.17  & 99.97 \\
    TrajHF \cite{li2025finetuning}              & \textbf{93.95} & \textbf{99.30} & 97.51  & 91.72     & \textbf{90.39}    & 98.02  & 99.81 \\ 
    \midrule
    AutoVLA (One-shot)                          & 80.54  &    96.89       &  92.42      &     94.43      &     75.82     &  88.06      &    99.94    \\
    AutoVLA (Post-RFT)                           &  89.11 &    98.41    &    95.64   &   95.40      &   81.87     &   \textbf{98.04}    &   99.94   \\
    AutoVLA (Best-of-N)                         & 92.12  & 99.14     & 97.08  & 95.51     & 87.55    & 97.12  & 99.98 \\
    \bottomrule
  \end{tabular}
  \vspace{-0.3cm}
\end{table}

\section{Experiments}
\label{experiments}
\subsection{Experimental Setup}

\textbf{Datasets.}
We train the AutoVLA model using a diverse set of real-world and simulation datasets. The nuPlan (Open-Scene) dataset \cite{karnchanachari2024towards,openscene2023} contains 120 hours of large-scale driving data with eight streams of camera data and object annotations. The Waymo end-to-end driving dataset (Waymo E2E) \cite{xu2025wod} comprises 4,021 20-second driving segments with eight streams of camera views and ego vehicle trajectories, especially focusing on challenging and long-tail scenarios, such as driving through construction areas or risky situations. The nuScenes dataset \cite{caesar2020nuscenes} provides 1,000 urban driving scenes with six camera views. The CARLA-Garage dataset \cite{Jaeger_2023_ICCV} provides over 500,000 frames of simulation camera data. In addition to the collected reasoning data, we utilize the DriveLM dataset \cite{sima2024drivelm} for nuScenes and CARLA datasets, by reformatting the VQA pairs to facilitate CoT reasoning.

\textbf{Benchmarks.}
We evaluate AutoVLA on both open-loop and closed-loop benchmarks across real-world and simulated environments. Open-loop performance is assessed on two public benchmarks: the NAVSIM benchmark \cite{dauner2024navsim} from the nuPlan dataset and the nuScenes benchmark \cite{hu2023planning}. The NAVSIM benchmark employs PDMS to assess key aspects of driving behavior, such as collision and ego progress. The nuScenes benchmark uses L2 distance and collision rate as evaluation metrics. Additionally, we report our model's performance on the Waymo end-to-end driving benchmark using the RFS metric, which reflects human-judged planning quality. Closed-loop performance is evaluated on the Bench2Drive benchmark \cite{jia2024bench2drive} in the CARLA simulator. Bench2Drive contains 44 interactive, closed-loop scenarios under varying locations and weather conditions, using metrics such as success rate, driving score, efficiency, and comfort. 

\begin{wrapfigure}{R}{0.5\linewidth}
    \centering
    \vspace{0.2cm}
    \includegraphics[width=0.5\textwidth]{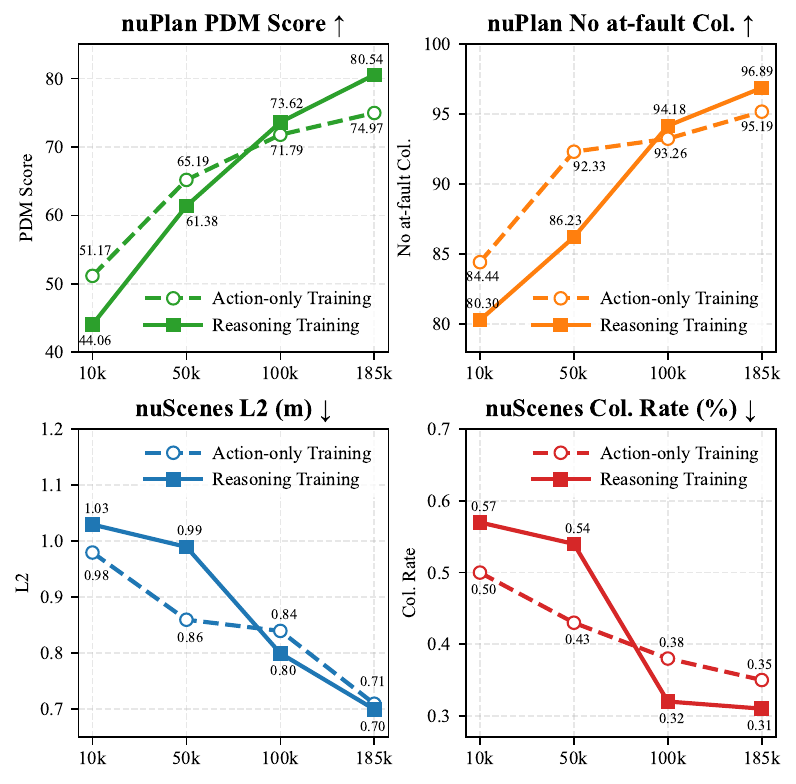}
    \caption{Data scaling effect on planning performance for nuPlan and nuScenes datasets (log-scaled x-axis). Increasing the amount of training data consistently enhances planning performance.}
    \label{fig:data_scaling}
    \vspace{-0.3cm}
\end{wrapfigure}

\textbf{Implementation Details.}
Each action token corresponds to 0.5 seconds of movement, and the planning horizon is set to 5 seconds. Consequently, the model outputs 10 action tokens, from which a 5-second trajectory can be decoded.
For SFT, we use a learning rate of $1 \times 10^{-5}$ and the FSDP training strategy. The model is trained for 5 epochs using 8 NVIDIA L40S GPUs. We use a per-GPU batch size of 1 and accumulate gradients over 4 steps, resulting in an effective batch size of 32. The weighting parameters in the SFT loss function are set to $\lambda_a = 1$ and $\lambda_{cot} = 40$.
For RFT, we employ the LoRA adapter \cite{hu2022lora} for parameter-efficient training. The learning rate for RFT is set to $3 \times 10^{-5}$, and the KL regularization weight $\beta$ is set to $0.04$. We perform a single policy update at each step, allowing the use of a simplified objective without the need for clipping or tracking the old policy. The model is fine-tuned for $6,000$ steps, and the best-performing checkpoint is selected for evaluation. Additional details are provided in the Supplementary Material.

\subsection{Main Results}

This section reports the main results of the AutoVLA model for various datasets and benchmarks, with additional results included in the Supplementary Material.

\textbf{Data Scaling Results.} 
We train AutoVLA on a mixture of the nuPlan and nuScenes datasets with varying training set sizes (10k, 50k, 100k, and the full 185k samples), with action-only supervision or with additional CoT reasoning supervision. The models are evaluated on the respective standard test sets, and the results are shown in \cref{fig:data_scaling}. We observe that increasing the amount of training data consistently improves planning performance on both datasets.
In the nuPlan dataset, when using fewer than 50k training samples, CoT reasoning does not outperform action-only in terms of PDMS and Collision Score. This is likely due to the increased difficulty of learning structured reasoning from limited data. However, as the training set size increases, models trained with CoT reasoning surpass those with action-only supervision, highlighting the scalability advantages of reasoning-augmented learning. A similar trend is observed on the nuScenes dataset: as the training set size increases, models trained with CoT reasoning consistently outperform those trained with action-only data in terms of L2 distance and collision rate.

\begin{wraptable}{r}{0.5\textwidth}
  \vspace{-0.3cm}
  \caption{Runtime Analysis of Fast \& Slow Thinking Modes in AutoVLA}
  \setlength{\tabcolsep}{6pt}
  \label{tab:thinking-modes}
  \centering
  \small
  \renewcommand{\arraystretch}{1.1}
  \begin{tabular}{l|ccc}
    \toprule
    Thinking Mode & Min. (s) & Max. (s) & Avg. (s) \\ 
    \midrule
    Fast Thinking & 0.997 & 1.116 & 1.072 \\
    Slow Thinking & 7.607 & 13.706 & 10.518 \\
    \bottomrule
  \end{tabular}
  \vspace{-0.2cm}
\end{wraptable}

\begin{figure}[t]
    \centering
    \includegraphics[width=1\linewidth]{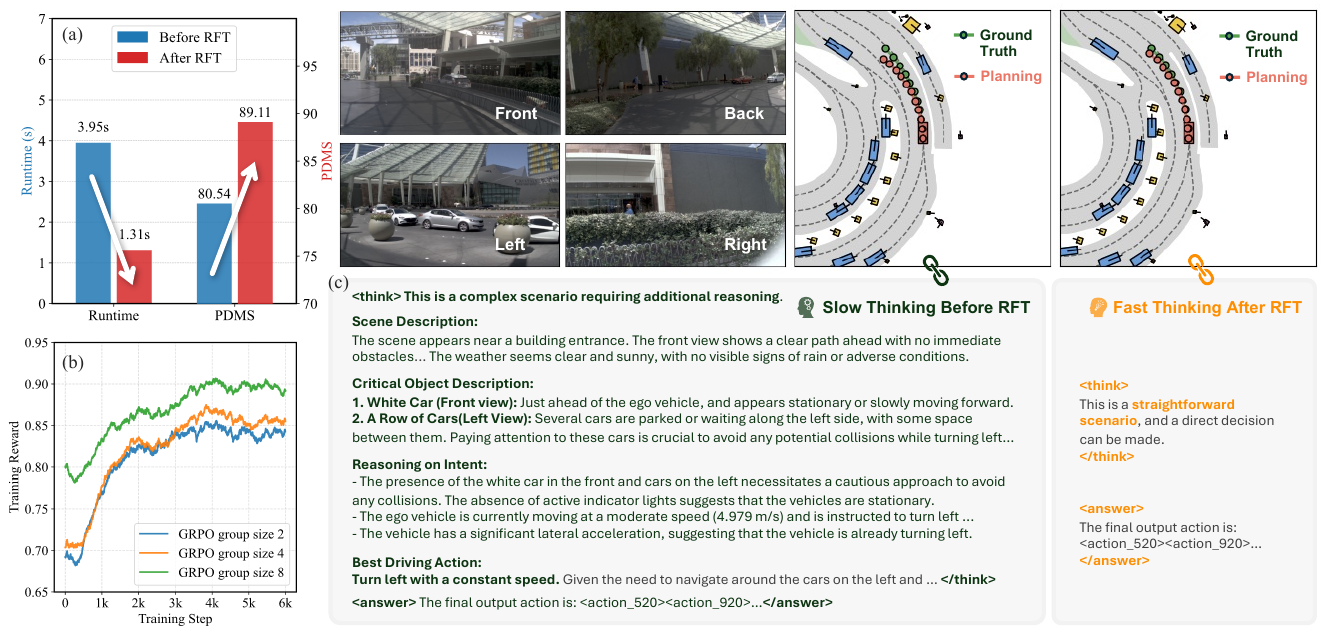}
    \vspace{-0.6cm}
    \caption{Reinforcement fine-tuning results on the nuPlan dataset. (a) Comparison of PDMS and runtime before and after RFT; (b) Training reward curves for different GRPO group sample sizes; (c) Qualitative comparison of planning and reasoning performance before and after RFT.}
    \label{fig:rft}
    \vspace{-0.4cm}
\end{figure}

\textbf{RFT Performance.} The slow-thinking mode incurs a significantly higher runtime due to the generation of CoT reasoning compared to the fast-thinking mode, as shown in \cref{tab:thinking-modes}. To mitigate this overhead, we introduce RFT to enhance AutoVLA’s adaptive thinking capability and avoid unnecessary reasoning in straightforward scenarios. Specifically, we apply RFT to the full-data CoT reasoning model trained via SFT to enhance its planning performance. As shown in \cref{fig:rft}(a), RFT yields a 10.6\% improvement in PDMS (on the NAVSIM testing set) and a 66.8\% reduction in runtime (average over 500 testing scenarios). The reward curve in \cref{fig:rft}(b) illustrates the progressive improvement of the model’s policy during RFT. Experiments with different GRPO group sample sizes indicate that larger groups lead to better performance by promoting broader exploration of training samples. As illustrated in \cref{fig:rft}(c), RFT also reduces unnecessary and slow reasoning in simple scenarios, driven by the CoT length penalty that encourages fast thinking for straightforward driving cases. A qualitative comparison shows that the SFT model produces suboptimal plans due to error accumulation in generation, whereas the RFT model (optimized via PDMS-based reward) generates better trajectories.

\textbf{nuPlan Benchmark Results.}
We evaluate AutoVLA against state-of-the-art end-to-end driving models on the NAVSIM benchmark \cite{dauner2024navsim} and present results in \cref{navsim}. In best-of-N planning, we use an oracle scorer to select the optimal trajectory from six generated candidates. After RFT, AutoVLA demonstrates significantly improved performance, aligning more closely with the NAVSIM reward signal. The best-of-N strategy further enhances performance, achieving the highest PDMS. Overall, AutoVLA achieves competitive results while demonstrating scalability across diverse datasets.

\begin{figure}[t]
    \centering
    \includegraphics[width=1\linewidth]{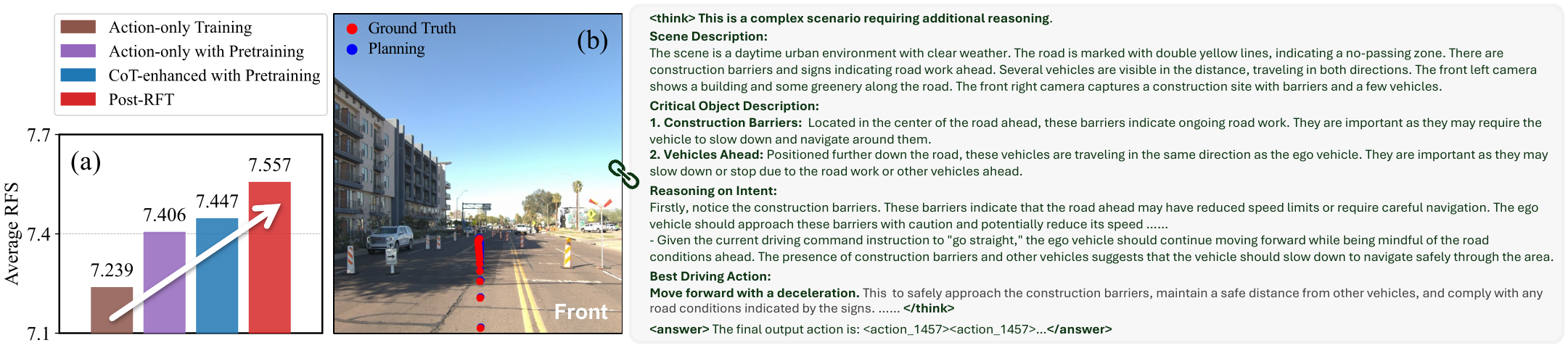}
    \vspace{-0.5cm}
    \caption{Performance comparison of AutoVLA with different training settings on the Waymo end-to-end driving dataset, along with an example illustrating the model's reasoning capabilities.}
    \label{fig:waymo}
    \vspace{-0.4cm}
\end{figure}

\textbf{Waymo E2E Performance.} 
We evaluate AutoVLA on the Waymo end-to-end driving dataset \cite{xu2025wod}, which features long-tail and complex driving scenarios. The model’s performance under various training settings on the test set is presented in \cref{fig:waymo}. The results reveal that pretraining on a combination of nuPlan and nuScenes datasets significantly enhances performance, suggesting enhanced scene understanding through exposure to more diverse training data. Incorporating CoT reasoning in training further improves planning performance compared to action-only training. Post-training with RFT, using ADE as the reward function, achieves the best overall RFS metric. A qualitative example in a construction zone demonstrates the model’s ability to reason about occlusions and generate effective detour plans.

\textbf{CARLA Closed-loop Performance.}
We evaluate the closed-loop driving performance of our AutoVLA model on the Bench2Drive benchmark \cite{jia2024bench2drive} in the CARLA simulator. The model is trained using SFT with both trajectory-only and CoT data. During testing, the planning frequency is set to 2 Hz. The results, shown in \cref{b2d}, demonstrate that AutoVLA outperforms existing end-to-end driving models in terms of overall driving score and success rate in the closed-loop test. 

\begin{table}[htp]
  \caption{Testing Results on the Bench2Drive (CARLA) Closed-loop Driving Benchmark}
  \vspace{0.2cm}
  \label{b2d}
  \centering
  \small
  \begin{tabular}{l|cccc}
    \toprule
    Method                                      & Driving Score $\uparrow$    &  Success Rate (\%) $\uparrow$  &  Efficiency $\uparrow$   & Comfortness $\uparrow$ \\
    \midrule
    AD-MLP \cite{zhai2023rethinking}            & 18.05           & 0.00                 &  48.45        & 22.63   \\
    UniAD-Base \cite{hu2023planning}            & 45.81           & 16.36                & 129.21        & 43.58   \\
    VAD \cite{jiang2023vad}                     & 42.35           & 15.00                & \textbf{157.94}        & \textbf{46.01}    \\
    TCP-traj \cite{wu2022trajectory}            & 59.90           & 30.00                & 76.54         & 18.08 \\
    DriveAdapter \cite{jia2023driveadapter}     & 64.22           & 33.08                & 70.22         & 16.01  \\
    Orion \cite{fu2025orion}                    & 77.74           & 54.62                & 151.48        & 17.38  \\ 
    \midrule
     AutoVLA                                    & \textbf{78.84}  & \textbf{57.73}      & 146.93        & 39.33 \\
    \bottomrule
  \end{tabular}
  \vspace{-0.4cm}
\end{table}

\subsection{Ablation Studies}

\begin{table}[t]
    \centering
    \small
    \renewcommand{\arraystretch}{1.08}
    \caption{Action Tokenization Accuracy with Different Codebook Sizes and Methods. (DCT: discrete
    cosine transform; MC: movement coverage; CU: codebook usage. \textbf{Bold} indicates best performance, \underline{underline} the second-best, and \uwave{tilde} the third-best.)}
    \label{tab:codebook size}
    \vspace{0.2cm}
    \setlength{\tabcolsep}{4pt}
    \begin{tabular}{c|cc|cc|cc|cc}
    \toprule
    \multirow[c]{2}{*}{\makecell{Codebook \\ Size}} & \multicolumn{2}{c|}{RT-1 (Action Bin) \cite{brohan2022rt}} & \multicolumn{2}{c|}{FAST (DCT) \cite{pertsch2025fast}} & \multicolumn{4}{c}{K-disk (Ours)}  \\ 
    \cmidrule(l){2-9}
                                   & ADE(m) $\downarrow$ & FDE(m) $\downarrow$ & ADE(m) $\downarrow$ & FDE(m) $\downarrow$ & ADE(m) $\downarrow$ & FDE(m) $\downarrow$ & MC $\uparrow$ & CU $\uparrow$ \\ 
    \midrule
    256                            & 0.1440  & 0.2942  & 0.1708  & 0.2137  & 0.0687  & 0.1034 &86.47\% & 100.0\%  \\
    1024                           & 0.1052  & 0.1883  & 0.0522  & 0.0588  & 0.0253  & 0.0282 &97.41\% & 100.0\%  \\
    2048                           & 0.1014  & 0.1775  & 0.0281  & 0.0309  & \uwave{0.0182}  & \uwave{0.0203} & \underline{99.42\%} & \textbf{100.0\%}  \\
    4096                           & 0.1001  & 0.1739  & \underline{0.0149}  & \underline{0.0161}  & \textbf{0.0141}  & \textbf{0.0155} & \textbf{100.0\%} & \underline{91.46\%} \\
    \bottomrule
    \end{tabular}
    \vspace{-0.4cm}
\end{table}

\textbf{Text Waypoint Output.} 
We use the same mixed training set from the nuPlan and nuScenes datasets to train a model that predicts waypoints in a text format, which are then converted into a trajectory of waypoints. We evaluate its performance in an open-loop planning setting using the standard test sets. The results, shown in \cref{text-waypoint}, indicate that our action tokenization and generation method significantly outperforms the text-based waypoint prediction approach. Additionally, due to the need to decode 
\begin{wraptable}{r}{0.5\textwidth}
    \centering
    \begin{minipage}{\linewidth}
        \centering
        \captionof{table}{Influence of Physical Action Tokenization}
        \label{text-waypoint}
        \setlength{\tabcolsep}{4pt}
        \small
        \begin{tabular}{l|cc}
            \toprule
            Metric                  & Text Waypoint & Physical Action \\
            \midrule
            PDM Score $\uparrow$      & 71.31         & 80.54 \\
            Avg. L2 (m) $\downarrow$  & 0.89          & 0.70 \\
            Avg. Col. (\%) $\downarrow$ & 0.36        & 0.31 \\ \midrule
            Runtime (s) $\downarrow$             & 7.65          & 3.95 \\
            \bottomrule
        \end{tabular}
    \end{minipage}

    \vspace{0.5cm} 

    \begin{minipage}{\linewidth} 
        \centering
        \captionof{table}{Performance Comparison across Different Tokenization Methods on NAVSIM}
        \label{tab:tokenization-compare}
        \vspace{0.2cm}
        \setlength{\tabcolsep}{3pt}
        \small
        \renewcommand{\arraystretch}{1.1}
        \begin{tabular}{l|ccc}
            \toprule
            Tokenization & PDMS $\uparrow$ & Collision $\uparrow$ & Progress $\uparrow$ \\
            \midrule
            FAST (DCT) \cite{pertsch2025fast} & 67.63 & 92.74 & 64.09 \\
            K-disk (Ours) & 80.54 & 96.89 & 75.82 \\
            \bottomrule
        \end{tabular}
    \end{minipage}
    
\end{wraptable}
numerical values, the text-based method incurs a substantially higher computational cost in generating the final trajectory. This shows the limitation of language models in handling precise numerical reasoning.

\textbf{Action Tokenization Methods.} 
We evaluate reconstruction accuracy across different codebook sizes for several tokenization methods, including RT-1 \cite{brohan2022rt}, FAST \cite{pertsch2025fast}, and our proposed K-disk tokenization, as shown in \cref{tab:codebook size}. For RT-1, acceleration and steering rate are discretized using uniform action bins, and a kinematic model is employed to reconstruct the trajectory. However, because only trajectory-level data is available and control actions must be indirectly inferred, this binning approach yields the highest reconstruction error. The FAST tokenization method is more sensitive to codebook size than our K-disk approach, achieving comparable reconstruction accuracy only at $K=4096$. In contrast, our proposed method consistently attains the highest reconstruction accuracy across all codebook sizes.

When the codebook size is small ($K=256$), reconstruction errors are significantly higher due to limited movement coverage, as many movements cannot be adequately represented by the small set of codebook tokens. As the codebook size increases beyond $K=2048$, the improvements in tokenization accuracy and movement coverage become marginal, while codebook usage decreases as many tokens remain redundant and unused. Considering this trade-off, we select $K=2048$ to balance reconstruction accuracy, movement coverage, and efficient codebook usage.

Furthermore, \cref{tab:tokenization-compare} reports planning performance on NAVSIM (nuPlan) for different tokenization methods. Our proposed method outperforms FAST. In FAST, the discrete cosine transform converts fixed-horizon planning trajectories into variable-length token sequences, complicating the selection of appropriate token lengths and making fixed-horizon planning difficult. 

\section{Conclusions}
\label{conclusions}

We propose AutoVLA, an end-to-end autonomous driving framework that unifies scene reasoning and action generation within a single autoregressive model. We adopt SFT to enable the model to operate in both fast thinking (direct trajectory generation) and slow thinking (enhanced with long CoT reasoning) modes. In addition, we introduce RFT to enable adaptive reasoning by penalizing unnecessary reasoning and aligning action generation with reward functions, improving both performance and efficiency. Experimental results demonstrate that AutoVLA achieves competitive performance on both open-loop and closed-loop planning benchmarks and exhibits strong reasoning capabilities. 

\textbf{Limitation and Future Work}. Although our model with dual-process adaptation achieves near-real-time inference (1 Hz), it remains highly GPU-dependent, requiring significant memory and computing. Future work will focus on real-time applications, optimizing runtime efficiency, and reducing computation overhead (e.g., through model quantization) to enable real-time deployment.

\section*{Acknowledgements}
This work was supported by the Federal Highway Administration Center of Excellence on New Mobility and Automated Vehicles, and by the National Science Foundation under Award No. 2346267, POSE: Phase II - DriveX: An Open Source Ecosystem for Automated Driving and Intelligent Transportation Research.

\bibliographystyle{unsrtnat}
\bibliography{ref}

\newpage 
\appendix
\input{supplementary}

\end{document}

%% file: supplementary.tex
\renewcommand{\thefigure}{S\arabic{figure}}
\renewcommand{\thetable}{S\arabic{table}}
\renewcommand{\theequation}{S\arabic{equation}}

\setcounter{figure}{0}
\setcounter{table}{0}
\setcounter{equation}{0}

\begin{center}
    {\LARGE \raisebox{-.15\height}{\includegraphics[width=0.065\textwidth]{figure/AutoVLA-Logo.png}} \hspace{0.1em} \textbf{\textit{AutoVLA} Supplementary Material}}
\end{center}

\section{Action Tokenization}

\subsection{Action Codebook}
\begin{figure}[h]
    \centering
    \includegraphics[width=\textwidth]{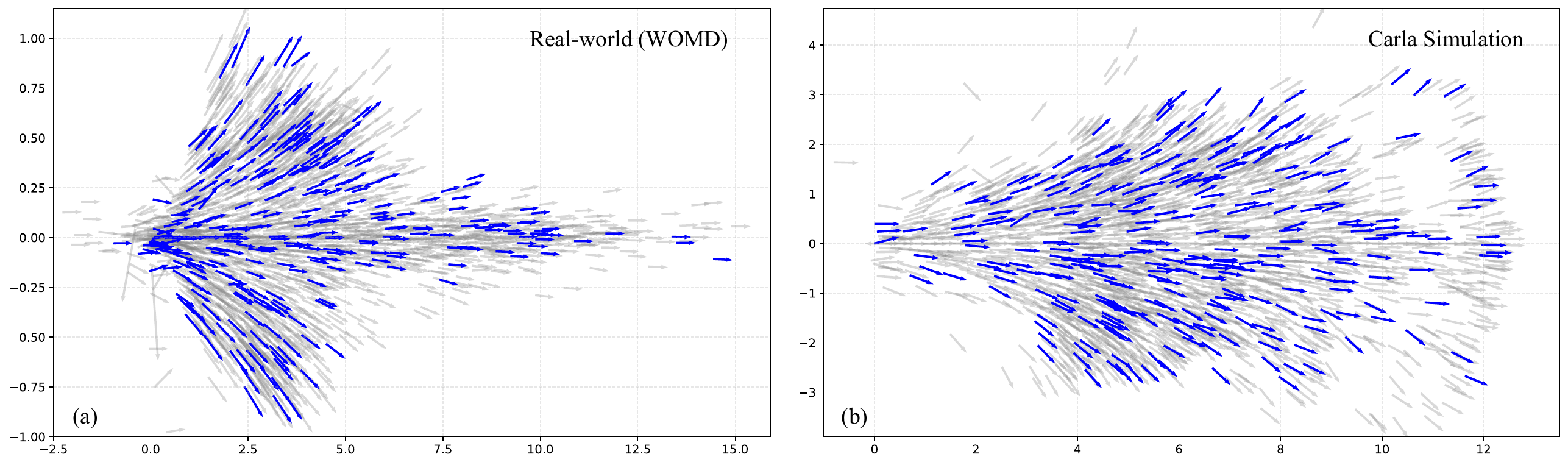}
    \caption{Visualization of action codebooks from (a) real-world dataset (WOMD) and (b) simulation dataset (CARLA). Grey arrows represent all 2048 tokens (position and heading), with 300 tokens randomly highlighted in blue for clarity.}
    \label{fig:action_token_cluster}
\end{figure}

To enable trajectory-level planning within a language model, we introduce \textbf{physical action tokens}, each representing a short-term feasible vehicle maneuver. These tokens are derived from clustering vehicle motion patterns in the Waymo Open Motion Dataset (WOMD) \cite{ettinger2021large}, which contains extensive real-world vehicle trajectories. The resulting codebook, visualized in \cref{fig:action_token_cluster}(a), is used across experiments involving multiple real-world datasets, including nuPlan, nuScenes, and Waymo. Due to differences in vehicle dynamics compared to real-world data for simulation testing, we construct a separate action codebook using the same clustering procedure on the CARLA-Garage dataset \cite{Jaeger_2023_ICCV}. This produces another set of 2048 action tokens for simulation testing, as shown in \cref{fig:action_token_cluster}(b).

We begin by sampling short motion segments from the dataset to construct a \textbf{discrete action codebook}. Each segment represents 0.5 seconds of vehicle motion, characterized by its final-frame bounding-box contour, computed based on the vehicle’s position, dimensions, and heading with respect to the current-frame coordinate. Then, we apply K-Disk clustering on these sampled segments, which iteratively selects a diverse set of representative segments $\{m_1, \dots, m_K\}$, such that no two segments are within a distance threshold $\delta = 0.05$ m, measured using average contour distance. For each selected segment $m_k$, we extract its spatial displacement and heading change, denoted as $(\Delta x, \Delta y, \Delta \theta)$, and define it as the action token $a_k$. The resulting action codebook is $\mathcal{A} = \{a_1, \dots, a_K\}$, with $K = 2048$, where each token encodes a distinct and physically feasible short-term vehicle behavior.

\subsection{Action Tokenizer} 
We implement an action tokenizer based on the constructed action codebook. During training, continuous trajectories are discretized by mapping each 0.5-second segment to its nearest action token in the codebook $\mathcal{A}$, resulting in a sequence of discrete tokens $[a_1, a_2, \dots, a_T]$.
During inference, the language model autoregressively generates a sequence of action tokens, each representing a 0.5-second motion segment. These tokens are then converted back to their corresponding motion segments in the codebook and applied sequentially from the initial ego pose. Therefore, the model reconstructs a continuous trajectory in the ego-centric coordinate frame by composing these local displacements and rotations.


\section{Reasoning Data Collection}

\begin{figure}[t]
    \centering
    \includegraphics[width=\linewidth]{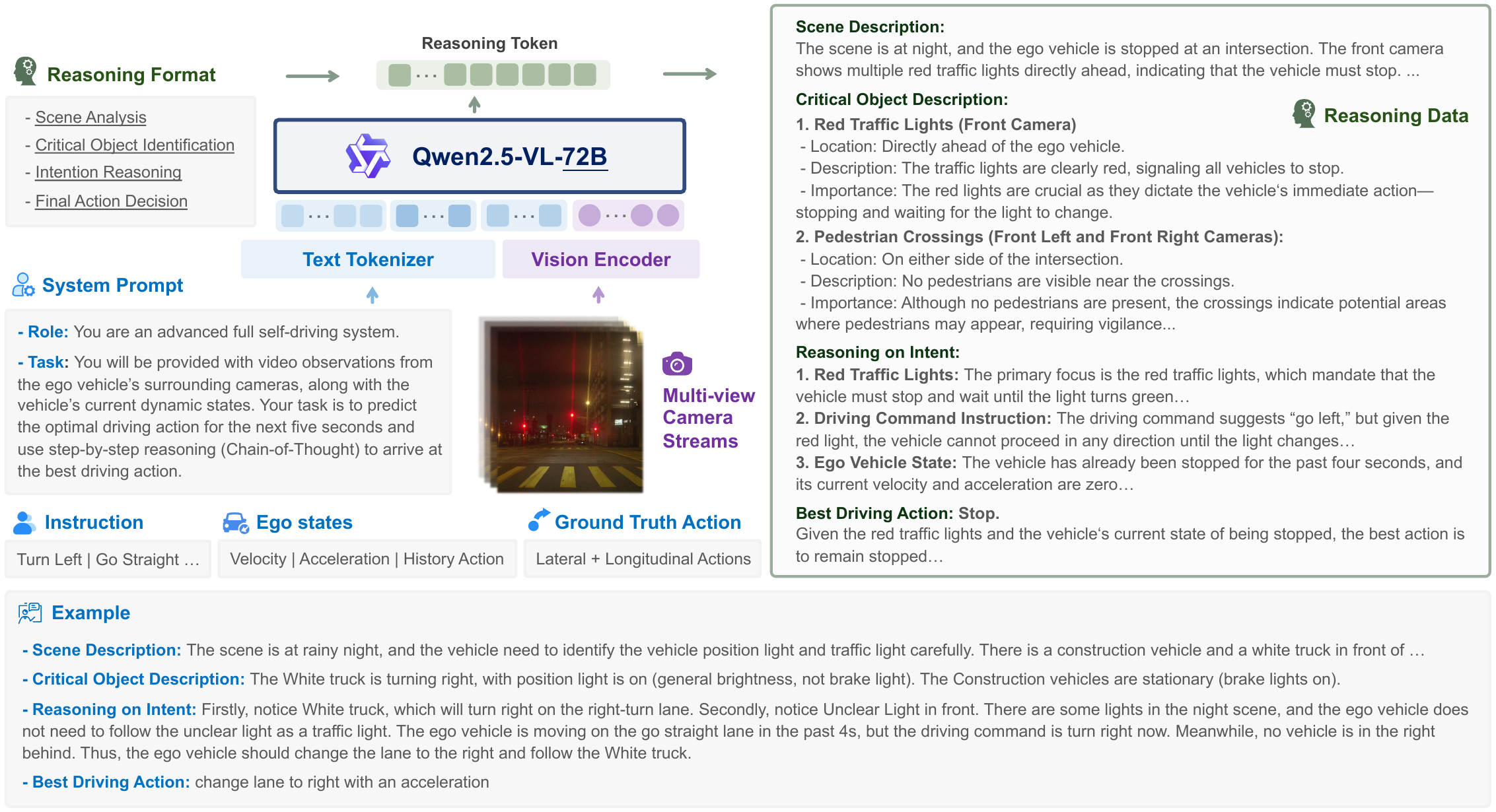}
    \caption{Automated reasoning annotation pipeline for autonomous driving, illustrated with a reasoning annotation example from the Waymo end-to-end driving dataset.}
    \label{fig:annotation}
\end{figure}

A large-scale, high-quality reasoning dataset with chain-of-thought (CoT) annotations is essential for enabling robust reasoning capabilities in vision-language-action (VLA) models. In this paper, we introduce an automated reasoning annotation pipeline using the state-of-the-art Qwen2.5-VL-72B vision-language model \cite{bai2025qwen2}, as illustrated in \cref{fig:annotation}. The pipeline significantly reduces reliance on human annotations and facilitates effective knowledge distillation from a more powerful, large-scale model to a more efficient, compact model. 

\subsection{Reasoning Annotation Pipeline} 
\textbf{System Prompt.} 
The system prompt specifies the model's role, task, expected CoT reasoning format, and examples of CoT reasoning. The definition of role and CoT reasoning format aligns with the required outputs of the AutoVLA model, which focuses on structured CoT reasoning. We carefully design several representative reasoning examples to guide the model. Moreover, the reasoning process includes four main steps: 1) scene description and analysis, 2) critical object identification and description, 3) intention reasoning of the surrounding objects, and 4) decision-making and meta-action.

\textbf{User Message.} 
The user message includes the driving instructions, ego vehicle states, and multi-view camera streams. Notably, we introduce the ground-truth driving meta-action derived from the data as explicit hints in reasoning, guiding the model to produce causal explanations that directly associate decisions with the driving context. This structured prompting significantly reduces nonsensical outputs and minimizes manual revisions.

\textbf{Reasoning Data Generation.} 
We employ the most capable model in the Qwen-VL series as the reasoning data annotation engine, leveraging its strong reasoning capacity and extensive world knowledge. The maximum length for newly generated reasoning textual tokens is set to 700. Additionally, we extract answers to relevant questions from the DriveLM VQA dataset \cite{sima2024drivelm}, which focuses on perception, prediction, and planning in the nuScenes and CARLA datasets, and then we reformat them into our standardized reasoning format. These reformatted samples are then combined with the generated data to construct the final reasoning dataset.

\textbf{Human Quality Check.}
We evaluate the quality of generated reasoning based on the accuracy and completeness of critical object identification, causal reasoning, and action decisions. The evaluation uses a binary scoring scheme: any error in these aspects results in a score of 0; otherwise, the sample receives a score of 1. Human annotators assessed 3,000 randomly selected samples, yielding an overall accuracy of 88.8\%, which demonstrates the high reliability of our proposed annotation pipeline. Erroneous samples were either corrected or discarded.






\subsection{Reasoning Annotation Visualization}
\begin{figure}[t]
    \centering
    \includegraphics[width=\linewidth]{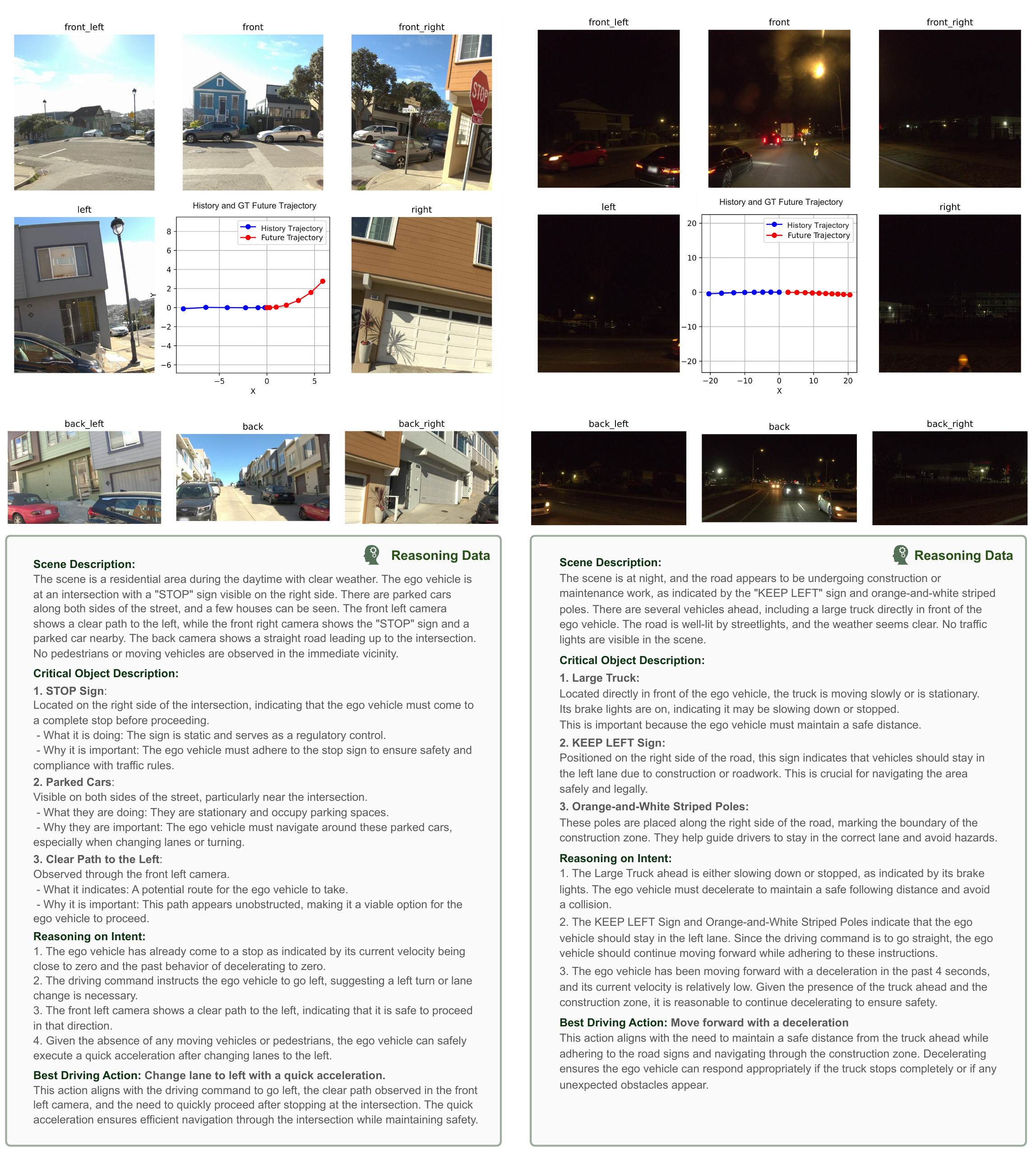}
    \caption{Visualization of the reasoning data annotation on the Waymo end-to-end driving dataset.}
    \label{fig:waymo_examples}
\end{figure}

\begin{figure}[t]
    \centering
    \includegraphics[width=\linewidth]{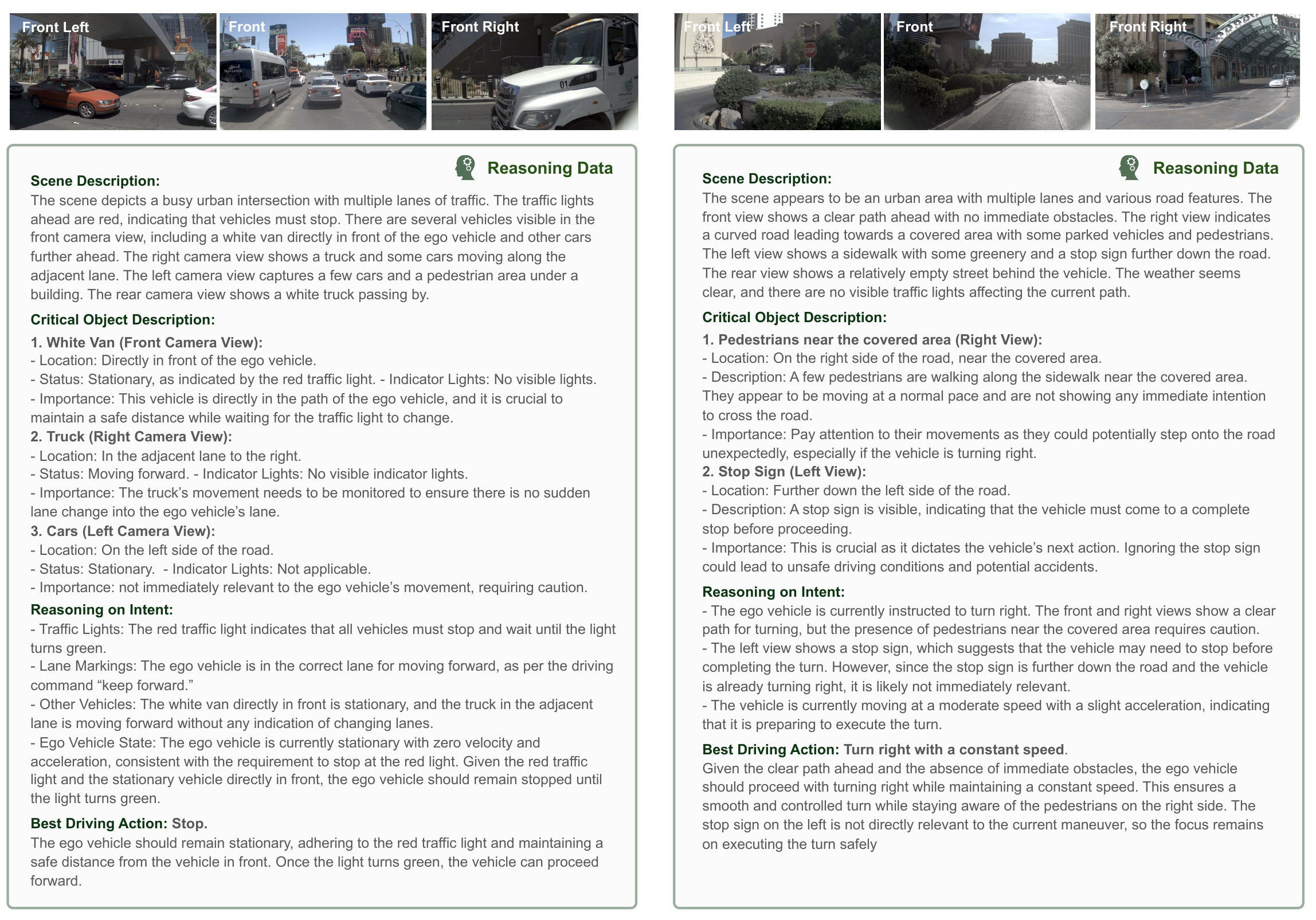}
    \caption{Visualization of the reasoning data annotation on the nuPlan dataset.}
    \label{fig:nuplan_examples}
\end{figure}

Some CoT reasoning annotation examples from the Waymo end-to-end driving dataset are shown in \cref{fig:waymo_examples}, showcasing that our annotation pipeline can deal with situations where a vehicle already stopped at a stop sign can proceed, rather than remaining stopped indefinitely. The pipeline also accurately interprets construction-related road control scenarios, enabling high-quality reasoning annotations. Examples on the nuPlan dataset are provided in \cref{fig:nuplan_examples}, demonstrating the pipeline’s capability to distinguish the functional relevance of stop signs and traffic lights across different lanes, rather than stopping at every detected stop sign.

\section{Details of Supervised Fine-tuning}

We perform supervised fine-tuning (SFT) with the Fully Sharded Data Parallel (FSDP) strategy for efficient multi-GPU training. We enable mixed-precision training using BFloat16 for parameters, gradients, and buffers to reduce memory usage and accelerate computation. Gradient checkpointing is enabled to reduce GPU memory consumption. The learning rate warm-up is 500 steps and decays by 2\% every 2,000 steps. The model is trained for 5 epochs, with gradient clipping applied at a maximum value of 1.0 to ensure training stability. 

\begin{figure}[t]
    \centering
    \includegraphics[width=\linewidth]{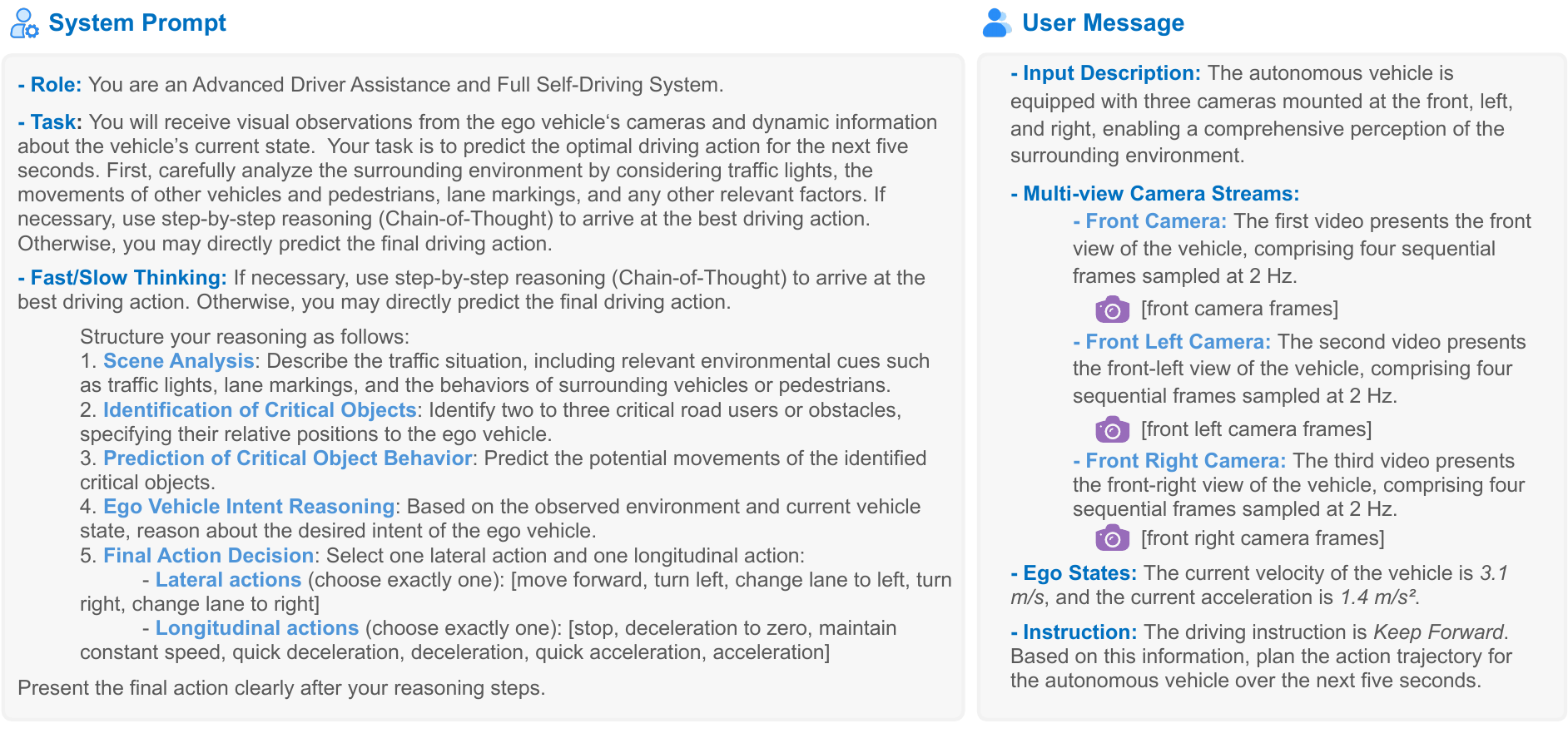}
    \caption{System Prompt and User Message of AutoVLA.}
    \label{fig:prompts}
    \vspace{-0.3cm}
\end{figure}

The model takes both system and user prompts as context inputs. As illustrated in \cref{fig:prompts}, the system prompt defines the model’s role, task, and the expected CoT reasoning format. The user message describes the multi-view camera observations, the ego vehicle’s current state (i.e., speed and acceleration) and optionally historical states, and the high-level driving instruction. We use three camera views (front, front-left, and front-right), each providing four consecutive frames. Images in the video stream are resized to maintain original aspect ratios but reduced to $28 \times 28 \times 128$ pixels.

\section{Details of Reinforcement Fine-tuning}

Group Relative Policy Optimization (GRPO) employs group-based sampling to compute the advantage function, replacing conventional state-value estimators or critic models. This design accelerates training while aligning naturally with the inherent multimodality of planning, which requires evaluating and selecting from a set of candidate trajectories. The overall reinforcement fine-tuning (RFT) procedure is illustrated in \cref{algo:grpo}.

\subsection{Kullback-Leibler (KL) Divergence}  
The KL divergence term $\mathbb{D}_{\text{KL}}$ is incorporated in \cref{grpo} to regularize the current policy with respect to the reference policy:
\begin{equation}
\mathbb{D}_{\text{KL}}(\pi_\theta \| \pi_{\text{ref}}) =
\frac{\pi_{\text{ref}}(o_i|q)}{\pi_\theta(o_i|q)} - 
\log \left( \frac{\pi_{\text{ref}}(o_i|q)}{\pi_\theta(o_i|q)} \right) - 1,
\end{equation}
where $\theta$ denotes the parameters of the current policy $\pi_{\theta}$, $q$ is the scenario input query, $o_i$ is the output of $i$th sample in the group, and $\pi_{\text{ref}}$ is the reference policy from the SFT stage. This regularization term penalizes large deviations from the reference policy, ensuring that policy updates remain within a stable area. As a result, the model retains useful knowledge acquired during SFT while improving driving behavior through reinforcement learning guided by validated reward functions.

\begin{algorithm}[t]
\caption{RFT for AutoVLA with GRPO}
\begin{spacing}{1.15}
\begin{algorithmic}[1]
\Require Supervised fine-tuned policy $\pi_{\mathrm{SFT}}$, Action Codebook $\mathcal{A}$, Group size $G$, Training step $K$, Dataset $D$, Driving reward function $r_\mathrm{Driving}$, Reasoning reward function $r_{\mathrm{CoT}}$, Balance weigh $\lambda_{r}$, KL regularization weight $\beta$
\Ensure Reinforcement fine-tuned policy $\pi_{\mathrm{RFT}}$
\State Initialize reference policy $\pi_{ref} \gets \pi_{\mathrm{SFT}}$ and current policy $\pi_{\theta} \gets \pi_{\mathrm{SFT}}$

\For{training step $1$ to $K$}
  \State Sample scenario $U$ from D
  \For{sample $i$ from $1$ to $G$}
    \State \parbox[t]{0.9\linewidth}{%
      Sample input query $q$, final output $o_i$, and per-token probability $\pi_{\theta}(o_i \vert q)$}
    \State $\pi_{\theta_{old}}(o_i \vert q)\gets  \pi_{\theta}(o_i \vert q)$
    \State Calculate per-token probability $\pi_{ref}(o_i|q)$ of $\pi_{ref}$ under $o_i$
    \State Decode trajectory $\tau \gets \mathcal{A}(o_i)$
    \State Calculate reward $r_{i} \gets r_\mathrm{Driving}\left(\tau, U\right) - \lambda_{\mathrm{r}} r_{\mathrm{CoT}}(o_i)$ 
  \EndFor
  \State Group average reward $\bar{r}, \ \sigma_r \gets \mathrm{mean}(r_1,\dots,r_G), \ \mathrm{std}(r_1,\dots,r_G)$
  
  \State Group relative advantage $A_i \gets (r_i - \bar{r}) / (\sigma_r)$
  
  
  

  \State $\mathcal{L}_{\text{RFT}} \gets -\sum_i^G\left( - \frac{\pi_\theta(o_i|q)}{\pi_{\theta_{\text{old}}}(o_i|q)} A_i + \beta \mathbb{D}_{\text{KL}}(\pi_\theta(o_i|q) \| \pi_{\text{ref}}(o_i|q)) \right)$

  \State Optimize $\pi_{\theta}$ according to $\mathcal{L}_{\text{RFT}}$
\EndFor
\State \Return Converged policy $\pi_{\mathrm{RFT}}$
\end{algorithmic}
\end{spacing}
\label{algo:grpo}
\end{algorithm}

\subsection{Reward Function}
The reward function provides the primary RFT training signal, guiding the optimization of policy updates. To further enhance comprehensive driving performance beyond imitating expert driving, we adopt a hybrid reward design that integrates the primary driving reward $r_{\text{Driving}}$ and a CoT penalty $r_{\text{CoT}}$ to discourage lengthy reasoning. For the NAVSIM Benchmark, we employ Predictive Driver Model Score (PDMS) \cite{dauner2024navsim} as the primary reward. For the Waymo end-to-end dataset, due to the limited annotation of RFS, we employ the (normalized) Average Displacement Error (ADE) as the driving reward. The final reward function, balanced by a weighting coefficient $\lambda_{r}$, is defined as:
\begin{equation}
r=r_{\text{Driving}}-\lambda_{r}r_{\text{CoT}}.
\end{equation}

In the nuPlan dataset, we adopt the PDMS score as the driving reward signal $r_{\text{Driving}}$, which serves as a comprehensive measure of driving quality. It is defined as follows:
\begin{equation}
\label{equ:PDMS}
r_{\text{Driving}}= \text{PDMS} = \text{NC} \times \text{DAC} \times \left( \frac{5\text{TTC} + 2\text{C} + 5\text{EP}}{12} \right),
\end{equation}
where the components include No at-fault Collision (NC), Drivable Area Compliance (DAC), Ego Progress (EP), Comfort (C), and Time-to-Collision (TTC). Detailed descriptions of each component can be found in \cite{dauner2024navsim}. Each sub-score is expressed as a percentage, and their weighted combination yields a normalized score within the range $[0,1]$, reflecting aspects of safety, comfort, and progress. For any scenario in which the planner fails (due to output errors), we assign a score of $0$.

For the Waymo dataset, we define the driving reward $r_{\text{Driving}}$ based on the ADE metric: 
\begin{equation}
\label{equ:ade}
r_{\text{Driving}}=\frac{\delta-ADE}{\kappa},\ \text{ADE} = \frac{1}{T}\sum_{t=1}^{T} \| \hat{\mathbf{y}}_t - \mathbf{y}_t \|_2,
\end{equation}
where $\delta$ denotes the maximum displacement error, and $\kappa$ is a scaling factor to normalize the reward. The planning trajectory $\hat{\mathbf{y}}$ is evaluated against the ground truth trajectory $\mathbf{y}$, and ADE is computed as the average L2 distances over $T$ time steps. 

To penalize excessively long reasoning chains, we introduce a CoT penalty term $r_{\text{CoT}}$, defined via a sigmoid function that normalizes the length of reasoning:
\begin{equation}
r_{\text{CoT}}=\frac{1}{1+e^{-(L-L_{tol})\gamma}},
\end{equation}
where $L$ denotes the length of the CoT reasoning, $L_{tol}$ is the tolerance threshold, and $\gamma$ is a scaling coefficient that controls the steepness of the penalty curve.

\subsection{Implementation Details} 
We use the \texttt{navtrain} split of the nuPlan dataset for RFT for the NAVSIM benchmark and the \texttt{validation} split of the Waymo end-to-end driving dataset for the Waymo benchmark. The vision encoder of pretrained AutoVLA is frozen, and the model is fine-tuned using Low-Rank Adaptation (LoRA) to reduce training costs and memory consumption. Specifically, both the LoRA rank and alpha are set to 8, with a dropout rate of 0.1. The pretrained SFT model is used as the reference policy during optimization. The hyperparameters $\gamma$, $L_{tol}$, and learning rate are set to $2\times10^{-3}$, $400$, and $3\times10^{-5}$, respectively. To ensure that the driving reward signal is dominant, the regularization weight is set to a relatively small value, $\lambda_{r} = 0.3$. We also set $\delta=2$, $\kappa=10$ in the RFT of the Waymo dataset. Moreover, we configure the generation parameters with a sampling temperature of $1.0$, top-p of $1.0$, and top-k of $0.0$ to encourage diverse and exploratory generation during GRPO sampling, which effectively covers a wider range of possible actions.

\section{Experiment Details}

\subsection{Data Preprocessing}

To enable mixed training across multiple driving datasets, we develop a unified data preprocessing pipeline that standardizes the format across all datasets. For each sample, we extract and standardize: 1) Ground truth trajectory coordinates and headings in the ego vehicle's coordinate frame at 2 Hz, 2) Image paths for multi-view camera image sequences consisting of 4 consecutive frames captured at 2 Hz (providing 2 seconds of history), 3) CoT reasoning annotations, 4) Vehicle states including its current velocity and acceleration, 5) High-level driving instructions. The preprocessing pipeline handles dataset-specific differences in data format, sampling rates, and coordinate systems to create a consistent format. The size and distribution of the final formatted dataset are shown in \cref{datasize}.

\textbf{nuPlan (NAVSIM).} 
We randomly sample 45.6k scenarios from the nuPlan \texttt{trainval} split and generate reasoning data using our proposed automated annotation pipeline. The resulting reasoning samples, together with the remaining training data with only trajectory annotation, constitute our full training set of nuPlan. Following the NAVSIM benchmark, the \texttt{navtest} split is used as the test set.

\textbf{nuScenes.} 
We preprocess all samples from the \texttt{training} set. For samples included in the DriveLM dataset, we reformat the question-answer (QA) pairs to generate structured reasoning annotations following our four-step reasoning format. Samples not covered by DriveLM are also used for training but only with trajectory supervision. The \texttt{validation} set is used for testing.

\textbf{Waymo.} 
The Waymo end-to-end driving dataset provides 2037 \texttt{training} and 479 \texttt{validation} segments, each containing a 20-second video with driving logs for the entire duration. We sample reasoning data using a 4-second sliding window and extract trajectory-only data using a 2-second sliding window offset by 1 second from the reasoning samples. Due to noise in the position data, the estimated vehicle heading can exhibit abrupt fluctuations when the vehicle is stationary. To address this issue, we apply a motion threshold to detect stationary periods and smooth the heading accordingly. The \texttt{test} set comprises 1505 samples.

\textbf{CARLA.} 
The CARLA-Garage dataset is used to train the model for closed-loop evaluation. Since only front camera images are available, we use single-view inputs (with four consecutive frames) for CARLA training and testing. We sample data using a sliding window with an offset of 0.5 seconds, and downsample the trajectories from 4 Hz to 2 Hz. For reasoning annotations, we leverage the DriveLM-CARLA dataset, which provides QA pairs similar to DriveLM-nuScenes, and we reformat the QA pairs to generate samples with reasoning annotations. 

\begin{table}[t]
  \caption{Training and Testing Data Size for Different Datasets}
  \vspace{0.2cm}
  \footnotesize
  \label{datasize}
  \centering
  \begin{tabular}{l|ccc}
    \toprule
    Dataset                                 & Train Samples     & Reasoning Samples   & Test Samples\\
    \midrule
    nuPlan (NAVSIM) \cite{dauner2024navsim} & 166.3k            &45.6k                  & 12.1k   \\
    nuScenes \cite{caesar2020nuscenes}      & 19.0k             & 2.9k                  & 5.6k \\
    Waymo \cite{xu2025wod}               & 23.8k             & 7.2k                  & 1.5k \\ 
    CARLA \cite{sima2024drivelm}            & 274.5k            & 53.2k                 & -- \\ 
    \bottomrule
  \end{tabular}
\end{table}

\subsection{Evaluation Metrics}

\textbf{nuPlan (NAVSIM).} 
\textbf{PDMS} is the official benchmark metric in NAVSIM on the nuPlan dataset, with its formulation provided in \cref{equ:PDMS}.

\textbf{nuScenes.} 
Following previous works, we adopt the \textbf{L2 Distance} to measure the average displacement error between predicted and ground-truth trajectories, and use \textbf{Collision Rate} to assess the frequency of predicted trajectories overlapping with surrounding objects. We follow the UniAD protocol, reporting both metrics at 1, 2, and 3 seconds in the future, instead of averaging over the horizons.

\textbf{Waymo.}
Following the Waymo end-to-end driving benchmark, \textbf{Rater Feedback Score (RFS)} is used to evaluate the driving performance. The Waymo dataset provides three human-annotated trajectories, each associated with a scalar quality score in $[0,10]$. A planning trajectory is matched to its closest reference trajectory, and a trust region is defined around it using lateral and longitudinal thresholds at fixed timesteps (3 s and 5 s). These thresholds are scaled by a piecewise linear function of the rater's trajectory’s initial speed. Predictions within the trust region inherit the matched rater's score, and those outside are penalized exponentially based on normalized deviation.tr

\textbf{CARLA.}
We evaluate driving performance in closed-loop testing using four key metrics: \textbf{Driving Score}, \textbf{Success Rate}, \textbf{Efficiency}, and \textbf{Comfortness} following the Bench2Drive benchmark \cite{dauner2024navsim}. Driving Score reflects the overall route completion penalized by infraction severity. Success Rate measures the percentage of routes completed without major infractions within the time limit. Efficiency quantifies how well the ego vehicle maintains a reasonable speed relative to surrounding traffic, computed as the average speed ratio over multiple checkpoints. Comfortness assesses the smoothness of the trajectory based on physical signals such as acceleration, jerk, and yaw rate, using expert-derived thresholds.

\subsection{Model Training}

Our primary base model is trained on the mixed training set of the nuPlan and nuScenes datasets. The ego state contains the vehicle's current velocity and acceleration. For the test on the Waymo end-to-end driving dataset, we further fine-tune this base model using the Waymo end-to-end driving dataset. For the model trained on Waymo, the ego vehicle's state encompasses current acceleration and a 4-second history of vehicle positions and velocities. For closed-loop simulation testing, we train a separate model using data preprocessed from the CARLA Garage dataset and DriveLM-CARLA annotations. This model is trained with single-view inputs instead of the multi-view setup used in the primary model, and we use a large resolution for the input images (with $28 \times 28 \times 384$ pixels).

\subsection{Inference}
We employ a stochastic generation strategy using top-p and top-k sampling to generate reasoning and planning outputs. Higher sampling diversity (e.g., temperature=$1.0$, top-p=$0.5$,  top-k=$20$) supports slow thinking modes, enabling the model to produce deeper and more elaborate reasoning chains. In contrast, more deterministic settings (e.g., temperature=$0.1$, top-p=$0.01$, top-k=$1$) produce fast thinking, yielding consistent and direct responses.

\begin{figure}[t]
    \centering
    \includegraphics[width=\linewidth]{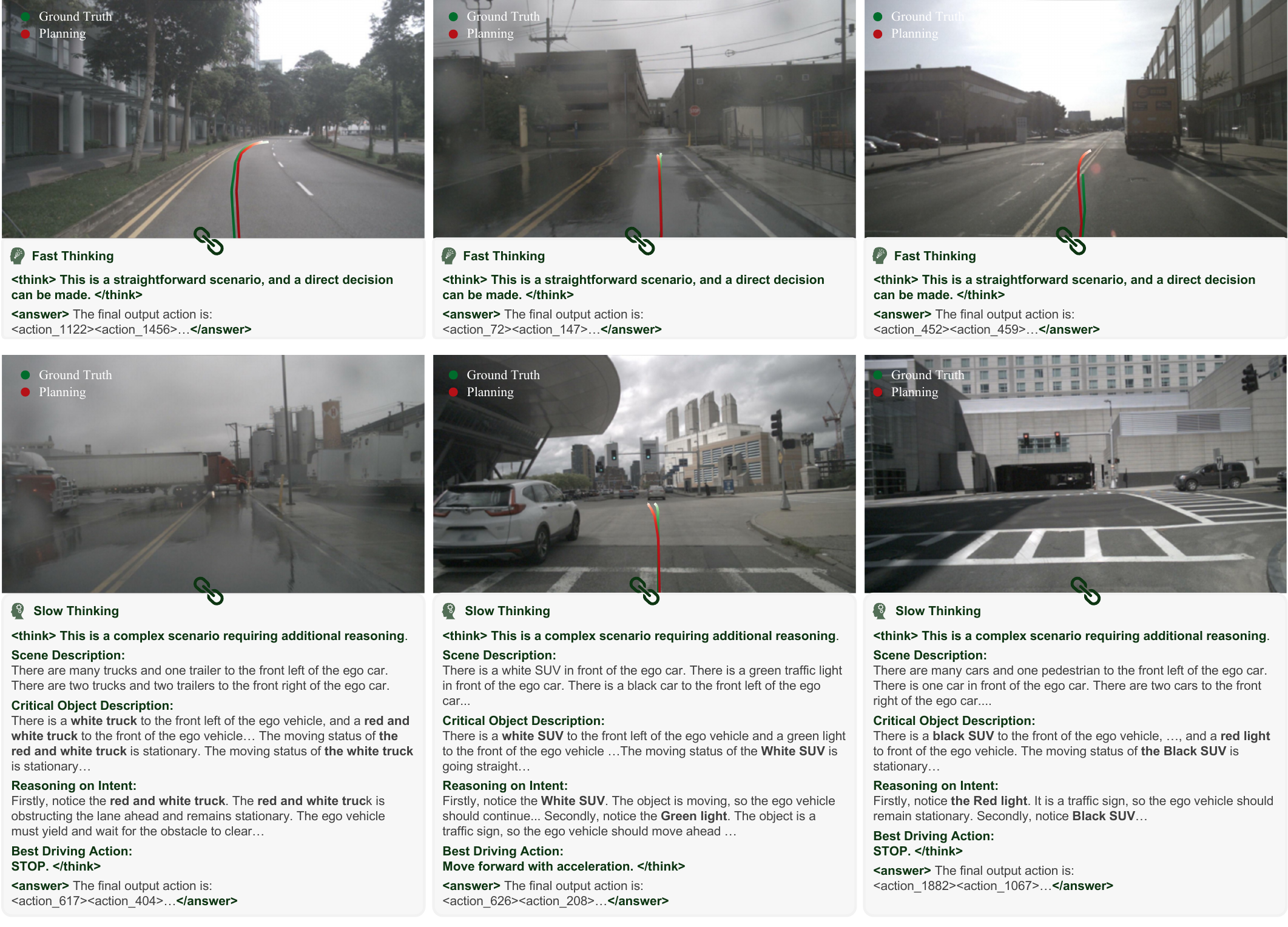}
    \caption{Planning and reasoning results of AutoVLA on the nuScenes dataset.}
    \label{fig:nusc_vis}
\end{figure}

\begin{table}[t]
  \caption{Testing Results of AutoVLA on the nuScenes Planning Benchmark}
  \vspace{0.2cm}
  \label{tab:nuscenes_quantitative}
  \centering
  \resizebox{\textwidth}{!}{
  \setlength{\tabcolsep}{3pt}
  \begin{tabular}{lcccccccccccccccc}
    \toprule
    \multirow{3}{*}{Method} 
    & \multicolumn{8}{c}{\textbf{ST-P3 metrics}} 
    & \multicolumn{8}{c}{\textbf{UniAD metrics}} \\
    \cmidrule(lr){2-9} \cmidrule(lr){10-17}
    & \multicolumn{4}{c}{L2 (m) ↓} 
    & \multicolumn{4}{c}{Collision (\%) ↓} 
    & \multicolumn{4}{c}{L2 (m) ↓} 
    & \multicolumn{4}{c}{Collision (\%) ↓} \\
    \cmidrule(lr){2-5} \cmidrule(lr){6-9} \cmidrule(lr){10-13} \cmidrule(lr){14-17}
    & 1s & 2s & 3s & Avg. 
    & 1s & 2s & 3s & Avg. 
    & 1s & 2s & 3s & Avg. 
    & 1s & 2s & 3s & Avg.\\
    \midrule

    ST-P3 \cite{hu2022st} & 1.33 & 2.11 & 2.90 & 2.11 & 0.23 & 0.62 & 1.27 & 0.71 & -& -& -& -& -& -& -& -\\
    VAD \cite{jiang2023vad}  & 0.17 & 0.34 & 0.60 & 0.37 & 0.07 & 0.10 & 0.24 & 0.14  & -& -& -& -& -& -& -& - \\
    UniAD \cite{hu2023planning} & 0.44 & 0.67 & 0.96 & 0.69 & 0.04 & 0.08 & 0.23 & 0.12 & 0.48 & 0.96 & 1.65 & 1.03 & 0.05 & \textbf{0.17} & 0.71 & 0.31\\
    EMMA \cite{hwang2024emma}& \textbf{0.14} & \textbf{0.29} & \textbf{0.54} & \textbf{0.32} & -& -& -& - & -& -& -& -& -& -& -& - \\
    OpenEMMA \cite{xing2025openemma} & 1.45 & 3.21 & 3.76 & 2.81 & -& -& -& - & -& -& -& -& -& -& -& - \\
    OpenDriveVLA-3B \cite{zhou2025opendrivevla} & \textbf{0.14} & 0.30 & 0.55 & 0.33 & \textbf{0.02} & \textbf{0.07} & \textbf{0.22} & \textbf{0.10} & \textbf{0.19} & \textbf{0.58} & 1.24 & \textbf{0.67} & \textbf{0.02} & 0.18 & 0.70 & \textbf{0.30}\\
    \midrule
    AutoVLA (action only) &0.22 & 0.39 &  0.61 &0.41 & 0.10 & 0.17 & 0.28 & 0.18 & 0.29 & 0.67 & 1.17 & 0.71 & 0.15 & 0.34 & 0.56 & 0.35    \\

     AutoVLA (w/ CoT) & 0.21 & 0.38 & 0.60 & 0.40 & 0.13 & 0.18 & 0.28 & 0.20  & 0.28 & 0.66 & \textbf{1.16} & 0.70 & 0.14 & 0.25 & \textbf{0.53} & 0.31\\ 
    \bottomrule
  \end{tabular}
  }
\end{table}

\section{Additional Results}

\begin{figure}[t]
    \centering
    \includegraphics[width=\linewidth]{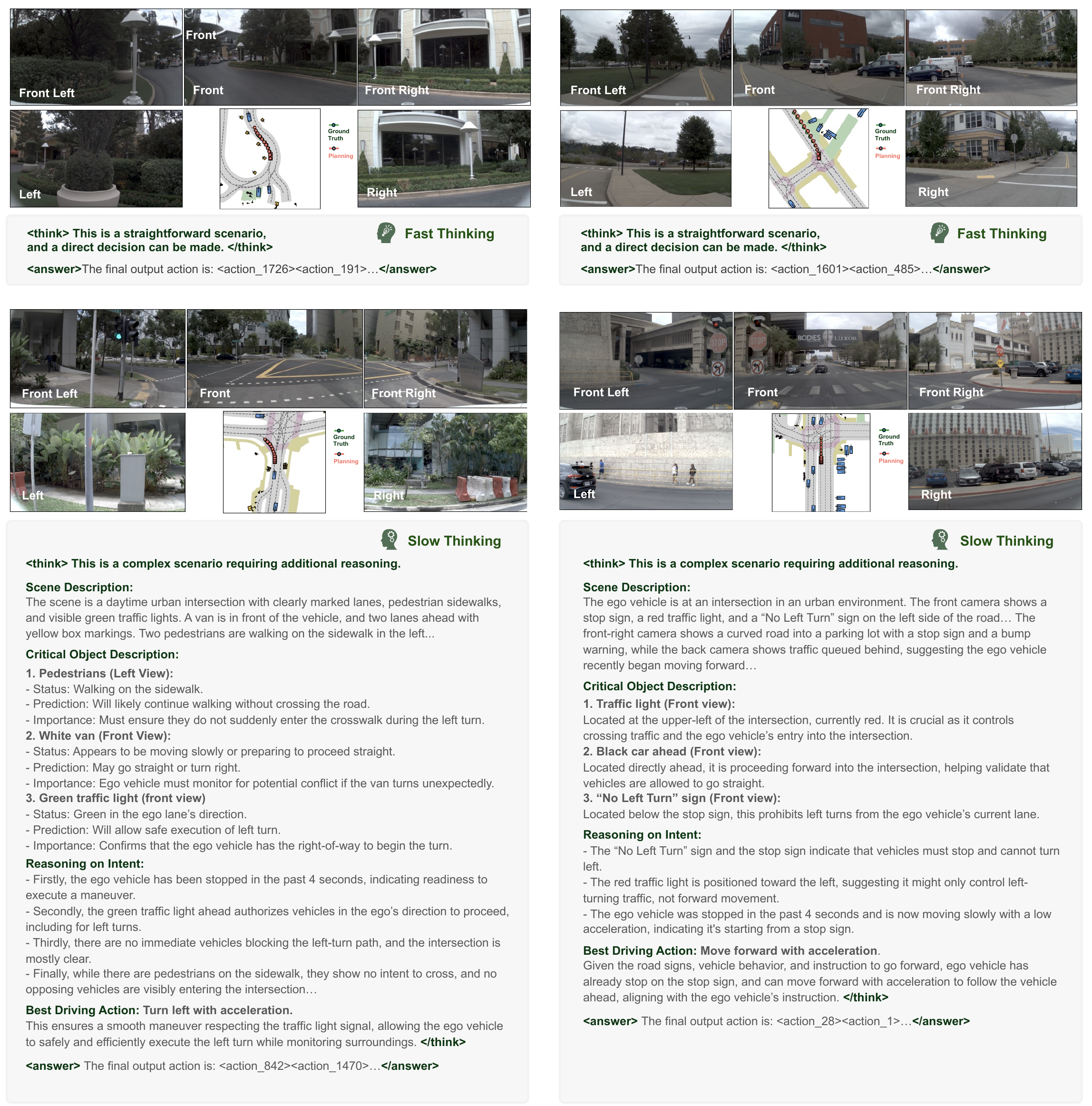}
    \caption{Planning and reasoning results of AutoVLA on the nuPlan dataset.}
    \label{fig:nuplan_vis}
\end{figure}

\begin{figure}[ht]
    \centering
    \includegraphics[width=\linewidth]{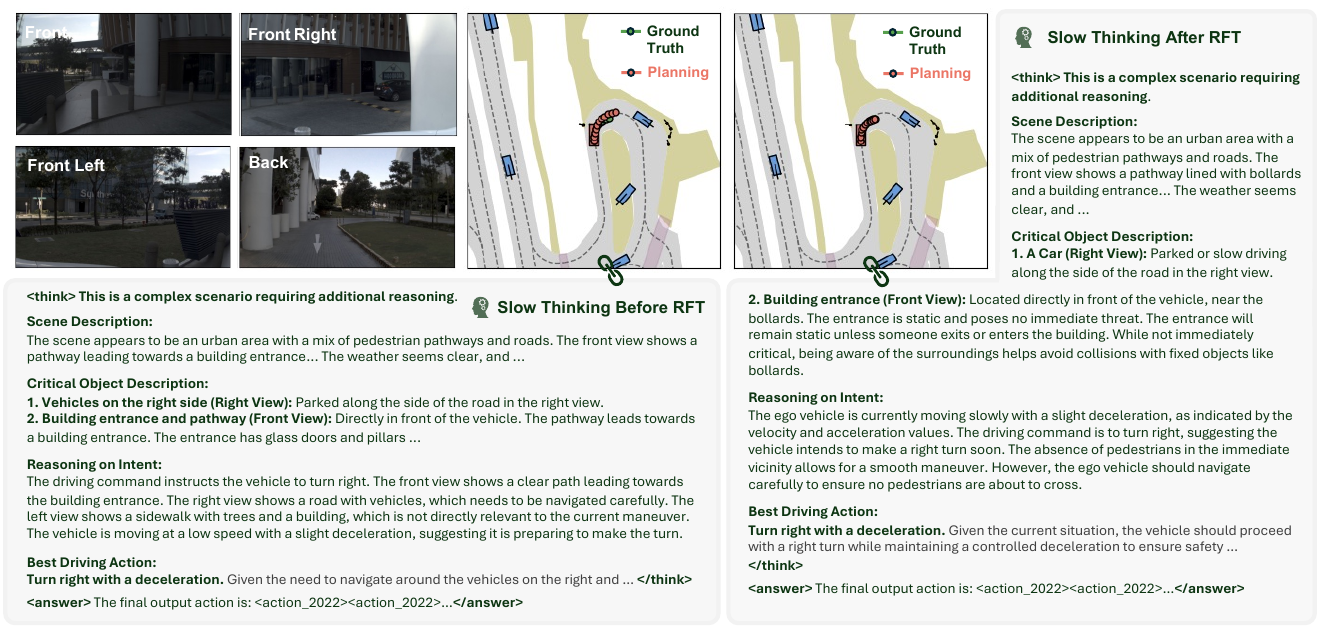}
    \caption{Qualitative comparison of planning and reasoning performance in complex scenarios, before and after RFT. Results indicate that RFT maintains reasoning capabilities in complex scenarios while enhancing planning performance.}
    \label{fig:keep slow thinking}
\end{figure}

\subsection{nuScenes Results}
We evaluate AutoVLA on the nuScenes dataset following both ST-P3 \cite{hu2022st} and UniAD \cite{hu2023planning} protocols, in comparison with state-of-the-art end-to-end models. As shown in \cref{tab:nuscenes_quantitative}, our model demonstrated competitive performance in the nuScenes planning benchmark. 

\cref{fig:nusc_vis} shows AutoVLA’s planning outputs. The model generates safe trajectories that closely align with ground-truth motions, accompanied by coherent and context-aware reasoning outputs. However, it can be observed that many nuScenes scenarios are relatively straightforward, often not requiring complex reasoning. This may explain the lack of performance gain in quantitative metrics when reasoning is introduced.

\subsection{nuPlan Results}
We present additional visualization results of our model on the nuPlan dataset in \cref{fig:nuplan_vis}. In relatively simple scenarios such as curved roads and intersections, our model generates high-quality trajectories via fast thinking. In more complex scenes with numerous traffic regulations, it leverages slow thinking to produce better and regulation-compliant planning results with CoT reasoning. As illustrated in \cref{fig:keep slow thinking}, our AutoVLA model preserves reasoning capabilities in complex scenarios despite removing redundant reasoning after RFT. Qualitative comparisons demonstrate that RFT enhances planning performance in complex scenarios.

\begin{figure}[ht]
    \centering
    \includegraphics[width=\linewidth]{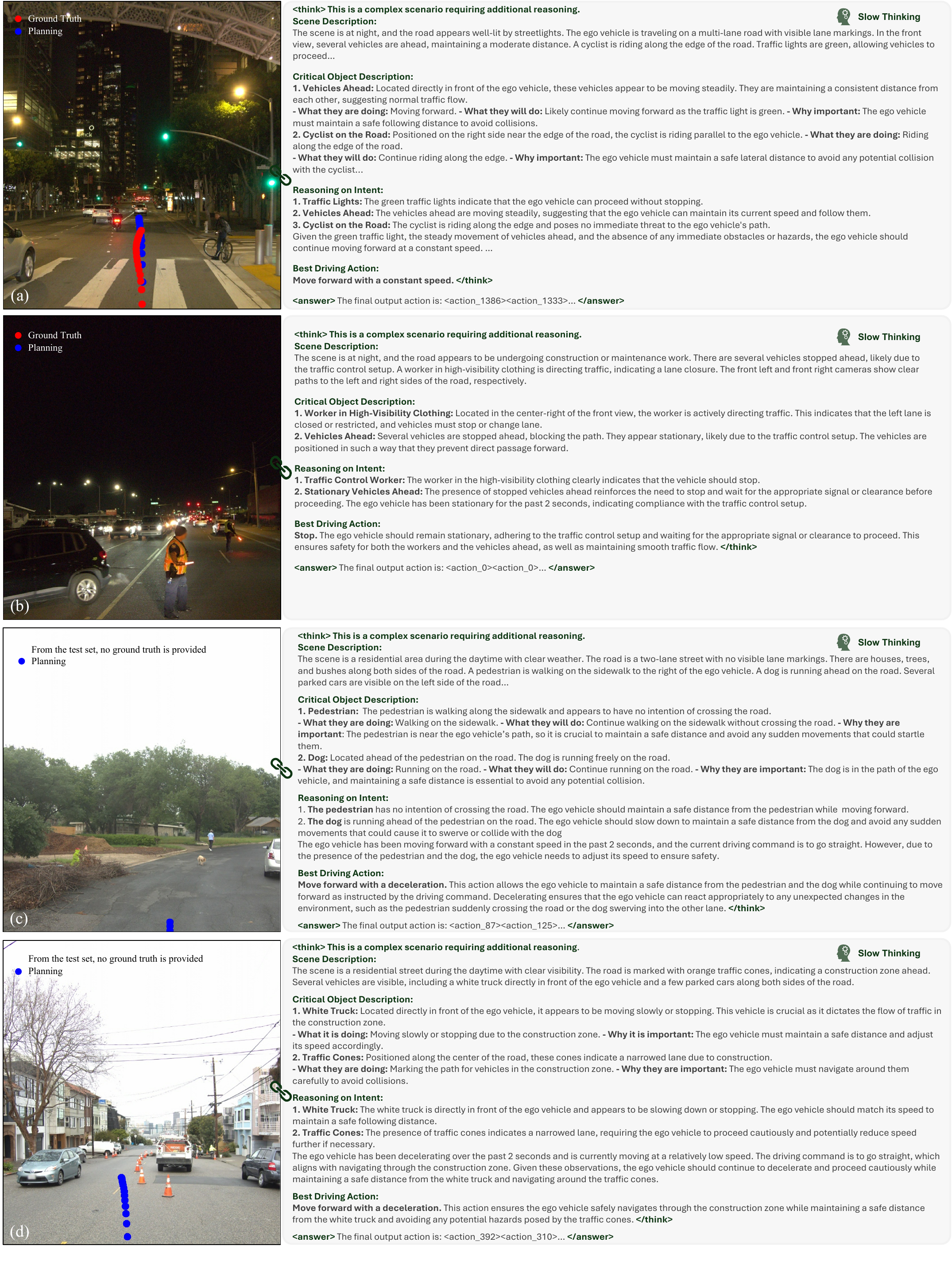}
    \caption{Planning and reasoning results of AutoVLA on the Waymo end-to-end driving dataset.}
    \label{fig:waymo_vis}
\end{figure}

\begin{table}[t]
\centering
\caption{Waymo Vision-based End-to-End Driving Challenge Leaderboard}
\vspace{0.2cm}
\renewcommand{\arraystretch}{1.15} 
\resizebox{\linewidth}{!}{
\begin{tabular}{l|c|c|c|c}
\toprule
\textbf{Method Name} & \textbf{RFS (Overall) $\uparrow$} & \textbf{ADE at 5s (Overall) $\downarrow$} & \textbf{ADE at 3s (Overall)$\downarrow$} & \textbf{RFS (Spotlight)$\uparrow$}\\
\midrule
Poutine & \textbf{7.9860} & 2.7419 & 1.2055 & 6.8929\\
HMVLM & 7.7367 & 3.0715 & 1.3269 & 6.7269\\
UniPlan & 7.6925 & 2.9864 & 1.3083 & 6.6544 \\
DiffusionLTF & 7.5919 & 2.9768 & 1.3605 &6.5688 \\
\underline{AutoVLA} & 7.5566 & 2.9580 & 1.3507 & \textbf{6.9436 }\\
Swin-Trajectory & 7.5432 & 2.8135 & 1.2082 & 6.6791 \\
waymo & 7.5281 & 3.0182 & 1.3200 &6.5953\\
ViT-Adapter-GRU & 7.4988 & \textbf{2.7024} & \textbf{1.1968} &6.4543\\
DriveTraj & 7.4957 & 2.9556 & 1.3038 &6.4101\\
open-llama & 7.4288 & 3.2165 & 1.3140 &6.2510\\
MTR-VP & 7.3433 & 3.3485 & 1.4232 &6.4023\\
WayPredict-XL & 7.2922 & 3.2915 & 1.4385 &6.3083\\
DriveTraj & 7.2787 & 3.4573 & 1.5346 &6.2428\\
WayPredict & 7.0641 & 3.5779 & 1.7242 &5.8562\\
LightEMMA & 6.5169 & 3.7395 & 1.7052 &5.7103\\
FrozenResNet50 & 6.4719 & 3.9148 & 1.9446 &5.7977\\
OpenEMMA & 5.1575 & 12.4755 & 6.6842 &4.7131\\
\bottomrule
\end{tabular}
}
\label{waymoe2e}
\end{table}

\begin{table}[t]
    \centering
    \caption{Ablation Study on the Waymo End-to-End Driving Test Set}
    \vspace{0.2cm}
    \footnotesize
    \begin{tabular}{l|l|l|cc}
    \toprule
    \textbf{Camera} & \textbf{Pretraining} & \textbf{Output}    & \textbf{RFS (Overall)} $\uparrow$ & \textbf{ADE at 5s} $\downarrow$ \\ \midrule
    Front           & None                 & Action-only        & 6.938                            & 3.595 \\
    Front           & None                 & CoT-enhanced       & 7.127                            & 3.188 \\
    Multi           & None                 & Action-only        & 7.239                            & 3.243 \\
    Multi           & None                 & CoT-enhanced       & 7.283                            & 3.182 \\
    Multi           & nuX                  & Action-only        & 7.406                            & 3.116 \\
    Multi           & nuX                  & CoT-enhanced       & 7.447                            & 3.115 \\ \midrule
    \multicolumn{3}{l|}{\textbf{Post-RFT}}                      & \textbf{7.557}                   & \textbf{2.958} \\
    \bottomrule
    \end{tabular}
    \label{tab:waymo_ablation}
\end{table}

\subsection{Waymo Results}

To overcome the limited size of the Waymo end-to-end driving dataset, we augment the training data with samples from the nuPlan and nuScenes datasets. For convenience, we refer to this combined dataset as \textbf{nuX}. We use this nuX data to pretrain the model before fine-tuning it on the Waymo dataset. To further enhance planning performance, we apply RFT on the AutoVLA model after SFT. Given the limited number of RFS-labeled samples (only 480 in the validation set), we adopt the average displacement error (ADE) as the primary reward signal.

As of May 22, 2025, AutoVLA achieves competitive performance on the Waymo End-to-End Driving Challenge leaderboard, as shown in \cref{waymoe2e}. The model ranks highly in both RFS Overall and ADE metrics and \textbf{achieves the top score in the RFS Spotlight metric}, which focuses on the most challenging scenarios. Qualitative results are shown in \cref{fig:waymo_vis}, where AutoVLA demonstrates its ability to generate safe, context-aware trajectories in complex environments. The model effectively handles interactions, construction zones, and traffic regulations in diverse scenarios while providing coherent reasoning to justify its decisions.


\cref{tab:waymo_ablation} presents the ablation studies of the AutoVLA model in different training setups. The results indicate that multi-camera input consistently enhances driving performance. When trained solely on the Waymo end-to-end driving dataset, incorporating reasoning significantly improves performance compared to action-only setups. Moreover, pretraining on nuX data provides a substantial performance boost, suggesting that such pretraining enhances the scene understanding of the model with more driving data. RFT can significantly improve planning performance by mitigating error accumulation in generation and aligning with the task-specific rewards.

\subsection{CARLA Results}
We evaluate the model after SFT in closed-loop testing within the CARLA simulator. The replanning frequency is set to 2 Hz, meaning the AutoVLA model is queried every 0.5 seconds in simulation, and the planned trajectory is used to generate control commands. High-level driving instructions are derived from a predefined route plan, while the vehicle’s current state (including speed and acceleration) is obtained from the IMU and speedometer sensors. The model receives four RGB images from the front-camera sensor covering the past two seconds as visual input. AutoVLA predicts a five-second trajectory, which is then used by a PID controller to compute the control actions (throttle, brake, and steering) that are applied to the vehicle.

\begin{figure}[ht]
    \centering
    \includegraphics[width=1\linewidth]{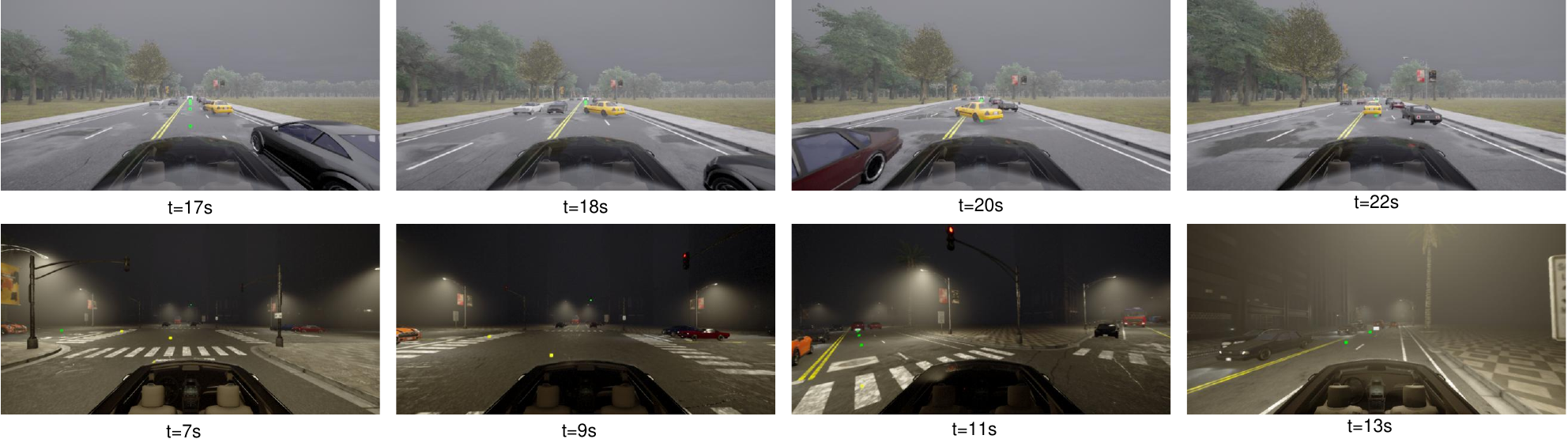}
    \caption{Closed-loop testing in the CARLA simulator. The two scenarios demonstrate AutoVLA's capability to (top) respond safely to a cut-in vehicle and (bottom) execute a smooth left turn.}
    \label{carla}
\end{figure}

Closed-loop testing results in the CARLA simulator are shown in \cref{carla}. Two representative scenarios are illustrated: (1) the ego vehicle equipped with AutoVLA successfully responds to a cut-in vehicle, and (2) it executes a smooth left turn. Additional closed-loop simulation results are available on the project website.

\section{Broader Impacts}
Autonomous driving is a safety-critical system, further emphasized by the integration of language guidance into the VLA model. This integration necessitates robust safeguards against adversarial attacks and proactive identification and filtering of unsafe human instructions. To mitigate potential hacking threats, it is essential to establish a secure communication channel in the vehicle, complemented by a gated-release strategy for model updates rather than online continual reinforcement fine-tuning on individual vehicles.


%% file: ref.bib
@article{kim2024openvla,
  title={Openvla: An open-source vision-language-action model},
  author={Kim, Moo Jin and Pertsch, Karl and Karamcheti, Siddharth and Xiao, Ted and Balakrishna, Ashwin and Nair, Suraj and Rafailov, Rafael and Foster, Ethan and Lam, Grace and Sanketi, Pannag and others},
  journal={arXiv preprint arXiv:2406.09246},
  year={2024}
}

@INPROCEEDINGS{li2024pretrain,
  author={Li, Yiheng and Zhao, Seth Z. and Xu, Chenfeng and Tang, Chen and Li, Chenran and Ding, Mingyu and Tomizuka, Masayoshi and Zhan, Wei},
  booktitle={2024 IEEE/RSJ International Conference on Intelligent Robots and Systems (IROS)}, 
  title={Pre-training on Synthetic Driving Data for Trajectory Prediction}, 
  year={2024},
  volume={},
  number={},
  pages={5910-5917},
  doi={10.1109/IROS58592.2024.10802492}
}

@article{bai2025qwen2,
  title={Qwen2.5-VL technical report},
  author={Bai, Shuai and Chen, Keqin and Liu, Xuejing and Wang, Jialin and Ge, Wenbin and Song, Sibo and Dang, Kai and Wang, Peng and Wang, Shijie and Tang, Jun and others},
  journal={arXiv preprint arXiv:2502.13923},
  year={2025}
}

@article{jiang2025alphadrive,
  title={Alphadrive: Unleashing the power of vlms in autonomous driving via reinforcement learning and reasoning},
  author={Jiang, Bo and Chen, Shaoyu and Zhang, Qian and Liu, Wenyu and Wang, Xinggang},
  journal={arXiv preprint arXiv:2503.07608},
  year={2025}
}

@inproceedings{weng2024drive,
  title={Para-drive: Parallelized architecture for real-time autonomous driving},
  author={Weng, Xinshuo and Ivanovic, Boris and Wang, Yan and Wang, Yue and Pavone, Marco},
  booktitle={Proceedings of the IEEE/CVF Conference on Computer Vision and Pattern Recognition},
  pages={15449--15458},
  year={2024}
}

@article{yuan2024drama,
  title={Drama: An efficient end-to-end motion planner for autonomous driving with mamba},
  author={Yuan, Chengran and Zhang, Zhanqi and Sun, Jiawei and Sun, Shuo and Huang, Zefan and Lee, Christina Dao Wen and Li, Dongen and Han, Yuhang and Wong, Anthony and Tee, Keng Peng and others},
  journal={arXiv preprint arXiv:2408.03601},
  year={2024}
}

@article{li2025finetuning,
  title={Finetuning Generative Trajectory Model with Reinforcement Learning from Human Feedback},
  author={Li, Derun and Ren, Jianwei and Wang, Yue and Wen, Xin and Li, Pengxiang and Xu, Leimeng and Zhan, Kun and Xia, Zhongpu and Jia, Peng and Lang, Xianpeng and others},
  journal={arXiv preprint arXiv:2503.10434},
  year={2025}
}

@inproceedings{arai2025covla,
  title={Covla: Comprehensive vision-language-action dataset for autonomous driving},
  author={Arai, Hidehisa and Miwa, Keita and Sasaki, Kento and Watanabe, Kohei and Yamaguchi, Yu and Aoki, Shunsuke and Yamamoto, Issei},
  booktitle={2025 IEEE/CVF Winter Conference on Applications of Computer Vision (WACV)},
  pages={1933--1943},
  year={2025},
  organization={IEEE}
}

@inproceedings{park2024vlaad,
  title={Vlaad: Vision and language assistant for autonomous driving},
  author={Park, SungYeon and Lee, MinJae and Kang, JiHyuk and Choi, Hahyeon and Park, Yoonah and Cho, Juhwan and Lee, Adam and Kim, DongKyu},
  booktitle={Proceedings of the IEEE/CVF Winter Conference on Applications of Computer Vision},
  pages={980--987},
  year={2024}
}

@article{zhang2024closed,
  title={Closed-Loop Supervised Fine-Tuning of Tokenized Traffic Models},
  author={Zhang, Zhejun and Karkus, Peter and Igl, Maximilian and Ding, Wenhao and Chen, Yuxiao and Ivanovic, Boris and Pavone, Marco},
  journal={arXiv preprint arXiv:2412.05334},
  year={2024}
}

@article{wu2024smart,
  title={SMART: scalable multi-agent real-time motion generation via next-token prediction},
  author={Wu, Wei and Feng, Xiaoxin and Gao, Ziyan and Kan, Yuheng},
  journal={Advances in Neural Information Processing Systems},
  volume={37},
  pages={114048--114071},
  year={2024}
}

@article{jia2024bench2drive,
  title={Bench2drive: Towards multi-ability benchmarking of closed-loop end-to-end autonomous driving},
  author={Jia, Xiaosong and Yang, Zhenjie and Li, Qifeng and Zhang, Zhiyuan and Yan, Junchi},
  journal={arXiv preprint arXiv:2406.03877},
  year={2024}
}

@article{fu2025orion,
  title={ORION: A Holistic End-to-End Autonomous Driving Framework by Vision-Language Instructed Action Generation},
  author={Fu, Haoyu and Zhang, Diankun and Zhao, Zongchuang and Cui, Jianfeng and Liang, Dingkang and Zhang, Chong and Zhang, Dingyuan and Xie, Hongwei and Wang, Bing and Bai, Xiang},
  journal={arXiv preprint arXiv:2503.19755},
  year={2025}
}

@article{li2024hydra,
  title={Hydra-mdp: End-to-end multimodal planning with multi-target hydra-distillation},
  author={Li, Zhenxin and Li, Kailin and Wang, Shihao and Lan, Shiyi and Yu, Zhiding and Ji, Yishen and Li, Zhiqi and Zhu, Ziyue and Kautz, Jan and Wu, Zuxuan and others},
  journal={arXiv preprint arXiv:2406.06978},
  year={2024}
}

@article{sima2025centaur,
  title={Centaur: Robust End-to-End Autonomous Driving with Test-Time Training},
  author={Sima, Chonghao and Chitta, Kashyap and Yu, Zhiding and Lan, Shiyi and Luo, Ping and Geiger, Andreas and Li, Hongyang and Alvarez, Jose M},
  journal={arXiv preprint arXiv:2503.11650},
  year={2025}
}

@article{huang2024gen,
  title={Gen-Drive: Enhancing Diffusion Generative Driving Policies with Reward Modeling and Reinforcement Learning Fine-tuning},
  author={Huang, Zhiyu and Weng, Xinshuo and Igl, Maximilian and Chen, Yuxiao and Cao, Yulong and Ivanovic, Boris and Pavone, Marco and Lv, Chen},
  journal={arXiv preprint arXiv:2410.05582},
  year={2024}
}

@article{chitta2022transfuser,
  title={Transfuser: Imitation with transformer-based sensor fusion for autonomous driving},
  author={Chitta, Kashyap and Prakash, Aditya and Jaeger, Bernhard and Yu, Zehao and Renz, Katrin and Geiger, Andreas},
  journal={IEEE transactions on pattern analysis and machine intelligence},
  volume={45},
  number={11},
  pages={12878--12895},
  year={2022},
  publisher={IEEE}
}

@inproceedings{hu2023planning,
  title={Planning-oriented autonomous driving},
  author={Hu, Yihan and Yang, Jiazhi and Chen, Li and Li, Keyu and Sima, Chonghao and Zhu, Xizhou and Chai, Siqi and Du, Senyao and Lin, Tianwei and Wang, Wenhai and others},
  booktitle={Proceedings of the IEEE/CVF conference on computer vision and pattern recognition},
  pages={17853--17862},
  year={2023}
}

@article{chen2024end,
  title={End-to-end autonomous driving: Challenges and frontiers},
  author={Chen, Li and Wu, Penghao and Chitta, Kashyap and Jaeger, Bernhard and Geiger, Andreas and Li, Hongyang},
  journal={IEEE Transactions on Pattern Analysis and Machine Intelligence},
  year={2024},
  publisher={IEEE}
}

@inproceedings{sima2024drivelm,
  title={Drivelm: Driving with graph visual question answering},
  author={Sima, Chonghao and Renz, Katrin and Chitta, Kashyap and Chen, Li and Zhang, Hanxue and Xie, Chengen and Bei{\ss}wenger, Jens and Luo, Ping and Geiger, Andreas and Li, Hongyang},
  booktitle={European Conference on Computer Vision},
  pages={256--274},
  year={2024},
  organization={Springer}
}

@article{mao2023gpt,
  title={Gpt-driver: Learning to drive with gpt},
  author={Mao, Jiageng and Qian, Yuxi and Ye, Junjie and Zhao, Hang and Wang, Yue},
  journal={arXiv preprint arXiv:2310.01415},
  year={2023}
}

@inproceedings{caesar2020nuscenes,
  title={nuscenes: A multimodal dataset for autonomous driving},
  author={Caesar, Holger and Bankiti, Varun and Lang, Alex H and Vora, Sourabh and Liong, Venice Erin and Xu, Qiang and Krishnan, Anush and Pan, Yu and Baldan, Giancarlo and Beijbom, Oscar},
  booktitle={Proceedings of the IEEE/CVF conference on computer vision and pattern recognition},
  pages={11621--11631},
  year={2020}
}

@article{zhang2024wisead,
  title={WiseAD: Knowledge Augmented End-to-End Autonomous Driving with Vision-Language Model},
  author={Zhang, Songyan and Huang, Wenhui and Gao, Zihui and Chen, Hao and Lv, Chen},
  journal={arXiv preprint arXiv:2412.09951},
  year={2024}
}

@article{sun2024sparsedrive,
  title={Sparsedrive: End-to-end autonomous driving via sparse scene representation},
  author={Sun, Wenchao and Lin, Xuewu and Shi, Yining and Zhang, Chuang and Wu, Haoran and Zheng, Sifa},
  journal={arXiv preprint arXiv:2405.19620},
  year={2024}
}

@inproceedings{ettinger2021large,
  title={Large scale interactive motion forecasting for autonomous driving: The waymo open motion dataset},
  author={Ettinger, Scott and Cheng, Shuyang and Caine, Benjamin and Liu, Chenxi and Zhao, Hang and Pradhan, Sabeek and Chai, Yuning and Sapp, Ben and Qi, Charles R and Zhou, Yin and others},
  booktitle={Proceedings of the IEEE/CVF International Conference on Computer Vision},
  pages={9710--9719},
  year={2021}
}

@article{zhai2023rethinking,
  title={Rethinking the open-loop evaluation of end-to-end autonomous driving in nuscenes},
  author={Zhai, Jiang-Tian and Feng, Ze and Du, Jinhao and Mao, Yongqiang and Liu, Jiang-Jiang and Tan, Zichang and Zhang, Yifu and Ye, Xiaoqing and Wang, Jingdong},
  journal={arXiv preprint arXiv:2305.10430},
  year={2023}
}

@InProceedings{Jaeger_2023_ICCV,
    author    = {Jaeger, Bernhard and Chitta, Kashyap and Geiger, Andreas},
    title     = {Hidden Biases of End-to-End Driving Models},
    booktitle = {Proceedings of the IEEE/CVF International Conference on Computer Vision (ICCV)},
    month     = {October},
    year      = {2023},
    pages     = {8240-8249}
}

@article{pertsch2025fast,
  title={Fast: Efficient action tokenization for vision-language-action models},
  author={Pertsch, Karl and Stachowicz, Kyle and Ichter, Brian and Driess, Danny and Nair, Suraj and Vuong, Quan and Mees, Oier and Finn, Chelsea and Levine, Sergey},
  journal={arXiv preprint arXiv:2501.09747},
  year={2025}
}

@article{zhao2025cot,
  title={CoT-VLA: Visual Chain-of-Thought Reasoning for Vision-Language-Action Models},
  author={Zhao, Qingqing and Lu, Yao and Kim, Moo Jin and Fu, Zipeng and Zhang, Zhuoyang and Wu, Yecheng and Li, Zhaoshuo and Ma, Qianli and Han, Song and Finn, Chelsea and others},
  journal={arXiv preprint arXiv:2503.22020},
  year={2025}
}

@inproceedings{karnchanachari2024towards,
  title={Towards learning-based planning: The nuPlan benchmark for real-world autonomous driving},
  author={Karnchanachari, Napat and Geromichalos, Dimitris and Tan, Kok Seang and Li, Nanxiang and Eriksen, Christopher and Yaghoubi, Shakiba and Mehdipour, Noushin and Bernasconi, Gianmarco and Fong, Whye Kit and Guo, Yiluan and others},
  booktitle={2024 IEEE International Conference on Robotics and Automation (ICRA)},
  pages={629--636},
  year={2024},
  organization={IEEE}
}

@article{jiang2024senna,
  title={Senna: Bridging large vision-language models and end-to-end autonomous driving},
  author={Jiang, Bo and Chen, Shaoyu and Liao, Bencheng and Zhang, Xingyu and Yin, Wei and Zhang, Qian and Huang, Chang and Liu, Wenyu and Wang, Xinggang},
  journal={arXiv preprint arXiv:2410.22313},
  year={2024}
}

@article{xu2024vlm,
  title={Vlm-ad: End-to-end autonomous driving through vision-language model supervision},
  author={Xu, Yi and Hu, Yuxin and Zhang, Zaiwei and Meyer, Gregory P and Mustikovela, Siva Karthik and Srinivasa, Siddhartha and Wolff, Eric M and Huang, Xin},
  journal={arXiv preprint arXiv:2412.14446},
  year={2024}
}

@article{wu2022trajectory,
  title={Trajectory-guided control prediction for end-to-end autonomous driving: A simple yet strong baseline},
  author={Wu, Penghao and Jia, Xiaosong and Chen, Li and Yan, Junchi and Li, Hongyang and Qiao, Yu},
  journal={Advances in Neural Information Processing Systems},
  volume={35},
  pages={6119--6132},
  year={2022}
}

@inproceedings{jia2023driveadapter,
  title={Driveadapter: Breaking the coupling barrier of perception and planning in end-to-end autonomous driving},
  author={Jia, Xiaosong and Gao, Yulu and Chen, Li and Yan, Junchi and Liu, Patrick Langechuan and Li, Hongyang},
  booktitle={Proceedings of the IEEE/CVF International Conference on Computer Vision},
  pages={7953--7963},
  year={2023}
}

@article{shao2024deepseekmath,
  title={Deepseekmath: Pushing the limits of mathematical reasoning in open language models},
  author={Shao, Zhihong and Wang, Peiyi and Zhu, Qihao and Xu, Runxin and Song, Junxiao and Bi, Xiao and Zhang, Haowei and Zhang, Mingchuan and Li, YK and Wu, Y and others},
  journal={arXiv preprint arXiv:2402.03300},
  year={2024}
}

@misc{openscene2023,
      title = {OpenScene: The Largest Up-to-Date 3D Occupancy Prediction Benchmark in Autonomous Driving},
      author = {OpenScene Contributors},
      howpublished={\url{https://github.com/OpenDriveLab/OpenScene}},
      year = {2023}
}

@article{philion2023trajeglish,
  title={Trajeglish: Traffic modeling as next-token prediction},
  author={Philion, Jonah and Peng, Xue Bin and Fidler, Sanja},
  journal={arXiv preprint arXiv:2312.04535},
  year={2023}
}

@article{wang2024drivecot,
  title={Drivecot: Integrating chain-of-thought reasoning with end-to-end driving},
  author={Wang, Tianqi and Xie, Enze and Chu, Ruihang and Li, Zhenguo and Luo, Ping},
  journal={arXiv preprint arXiv:2403.16996},
  year={2024}
}

@inproceedings{li2024driving,
  title={Driving everywhere with large language model policy adaptation},
  author={Li, Boyi and Wang, Yue and Mao, Jiageng and Ivanovic, Boris and Veer, Sushant and Leung, Karen and Pavone, Marco},
  booktitle={Proceedings of the IEEE/CVF Conference on Computer Vision and Pattern Recognition},
  pages={14948--14957},
  year={2024}
}

@article{xu2024drivegpt4,
  title={Drivegpt4: Interpretable end-to-end autonomous driving via large language model},
  author={Xu, Zhenhua and Zhang, Yujia and Xie, Enze and Zhao, Zhen and Guo, Yong and Wong, Kwan-Yee K and Li, Zhenguo and Zhao, Hengshuang},
  journal={IEEE Robotics and Automation Letters},
  year={2024},
  publisher={IEEE}
}

@inproceedings{chen2025automated,
  title={Automated evaluation of large vision-language models on self-driving corner cases},
  author={Chen, Kai and Li, Yanze and Zhang, Wenhua and Liu, Yanxin and Li, Pengxiang and Gao, Ruiyuan and Hong, Lanqing and Tian, Meng and Zhao, Xinhai and Li, Zhenguo and others},
  booktitle={2025 IEEE/CVF Winter Conference on Applications of Computer Vision (WACV)},
  pages={7817--7826},
  year={2025},
  organization={IEEE}
}

@article{renz2024carllava,
  title={CarLLaVA: Vision language models for camera-only closed-loop driving},
  author={Renz, Katrin and Chen, Long and Marcu, Ana-Maria and H{\"u}nermann, Jan and Hanotte, Benoit and Karnsund, Alice and Shotton, Jamie and Arani, Elahe and Sinavski, Oleg},
  journal={arXiv preprint arXiv:2406.10165},
  year={2024}
}

@article{gao2025rad,
  title={Rad: Training an end-to-end driving policy via large-scale 3dgs-based reinforcement learning},
  author={Gao, Hao and Chen, Shaoyu and Jiang, Bo and Liao, Bencheng and Shi, Yiang and Guo, Xiaoyang and Pu, Yuechuan and Yin, Haoran and Li, Xiangyu and Zhang, Xinbang and others},
  journal={arXiv preprint arXiv:2502.13144},
  year={2025}
}

@article{wang2024omnidrive,
  title={Omnidrive: A holistic llm-agent framework for autonomous driving with 3d perception, reasoning and planning},
  author={Wang, Shihao and Yu, Zhiding and Jiang, Xiaohui and Lan, Shiyi and Shi, Min and Chang, Nadine and Kautz, Jan and Li, Ying and Alvarez, Jose M},
  journal={arXiv preprint arXiv:2405.01533},
  year={2024}
}

@inproceedings{marcu2024lingoqa,
  title={LingoQA: Visual question answering for autonomous driving},
  author={Marcu, Ana-Maria and Chen, Long and H{\"u}nermann, Jan and Karnsund, Alice and Hanotte, Benoit and Chidananda, Prajwal and Nair, Saurabh and Badrinarayanan, Vijay and Kendall, Alex and Shotton, Jamie and others},
  booktitle={European Conference on Computer Vision},
  pages={252--269},
  year={2024},
  organization={Springer}
}

@article{tian2024drivevlm,
  title={Drivevlm: The convergence of autonomous driving and large vision-language models},
  author={Tian, Xiaoyu and Gu, Junru and Li, Bailin and Liu, Yicheng and Wang, Yang and Zhao, Zhiyong and Zhan, Kun and Jia, Peng and Lang, Xianpeng and Zhao, Hang},
  journal={arXiv preprint arXiv:2402.12289},
  year={2024}
}

@article{park2025nuplanqa,
  title={NuPlanQA: A Large-Scale Dataset and Benchmark for Multi-View Driving Scene Understanding in Multi-Modal Large Language Models},
  author={Park, Sung-Yeon and Cui, Can and Ma, Yunsheng and Moradipari, Ahmadreza and Gupta, Rohit and Han, Kyungtae and Wang, Ziran},
  journal={arXiv preprint arXiv:2503.12772},
  year={2025}
}

@article{renz2025simlingo,
  title={SimLingo: Vision-Only Closed-Loop Autonomous Driving with Language-Action Alignment},
  author={Renz, Katrin and Chen, Long and Arani, Elahe and Sinavski, Oleg},
  journal={arXiv preprint arXiv:2503.09594},
  year={2025}
}

@inproceedings{shao2024lmdrive,
  title={Lmdrive: Closed-loop end-to-end driving with large language models},
  author={Shao, Hao and Hu, Yuxuan and Wang, Letian and Song, Guanglu and Waslander, Steven L and Liu, Yu and Li, Hongsheng},
  booktitle={Proceedings of the IEEE/CVF Conference on Computer Vision and Pattern Recognition},
  pages={15120--15130},
  year={2024}
}

@article{zhou2025opendrivevla,
  title={OpenDriveVLA: Towards End-to-end Autonomous Driving with Large Vision Language Action Model},
  author={Zhou, Xingcheng and Han, Xuyuan and Yang, Feng and Ma, Yunpu and Knoll, Alois C},
  journal={arXiv preprint arXiv:2503.23463},
  year={2025}
}

@article{dauner2024navsim,
  title={Navsim: Data-driven non-reactive autonomous vehicle simulation and benchmarking},
  author={Dauner, Daniel and Hallgarten, Marcel and Li, Tianyu and Weng, Xinshuo and Huang, Zhiyu and Yang, Zetong and Li, Hongyang and Gilitschenski, Igor and Ivanovic, Boris and Pavone, Marco and others},
  journal={Advances in Neural Information Processing Systems},
  volume={37},
  pages={28706--28719},
  year={2024}
}

@article{cai2024driving,
  title={Driving with Regulation: Interpretable Decision-Making for Autonomous Vehicles with Retrieval-Augmented Reasoning via LLM},
  author={Cai, Tianhui and Liu, Yifan and Zhou, Zewei and Ma, Haoxuan and Zhao, Seth Z and Wu, Zhiwen and Ma, Jiaqi},
  journal={arXiv preprint arXiv:2410.04759},
  year={2024}
}

@article{hwang2024emma,
  title={Emma: End-to-end multimodal model for autonomous driving},
  author={Hwang, Jyh-Jing and Xu, Runsheng and Lin, Hubert and Hung, Wei-Chih and Ji, Jingwei and Choi, Kristy and Huang, Di and He, Tong and Covington, Paul and Sapp, Benjamin and others},
  journal={arXiv preprint arXiv:2410.23262},
  year={2024}
}

@inproceedings{xing2025openemma,
  title={Openemma: Open-source multimodal model for end-to-end autonomous driving},
  author={Xing, Shuo and Qian, Chengyuan and Wang, Yuping and Hua, Hongyuan and Tian, Kexin and Zhou, Yang and Tu, Zhengzhong},
  booktitle={Proceedings of the Winter Conference on Applications of Computer Vision},
  pages={1001--1009},
  year={2025}
}

@inproceedings{jiang2023vad,
  title={Vad: Vectorized scene representation for efficient autonomous driving},
  author={Jiang, Bo and Chen, Shaoyu and Xu, Qing and Liao, Bencheng and Chen, Jiajie and Zhou, Helong and Zhang, Qian and Liu, Wenyu and Huang, Chang and Wang, Xinggang},
  booktitle={Proceedings of the IEEE/CVF International Conference on Computer Vision},
  pages={8340--8350},
  year={2023}
}

@inproceedings{zheng2024genad,
  title={Genad: Generative end-to-end autonomous driving},
  author={Zheng, Wenzhao and Song, Ruiqi and Guo, Xianda and Zhang, Chenming and Chen, Long},
  booktitle={European Conference on Computer Vision},
  pages={87--104},
  year={2024},
  organization={Springer}
}

@article{liao2024diffusiondrive,
  title={DiffusionDrive: Truncated Diffusion Model for End-to-End Autonomous Driving},
  author={Liao, Bencheng and Chen, Shaoyu and Yin, Haoran and Jiang, Bo and Wang, Cheng and Yan, Sixu and Zhang, Xinbang and Li, Xiangyu and Zhang, Ying and Zhang, Qian and others},
  journal={arXiv preprint arXiv:2411.15139},
  year={2024}
}

@article{li2024bevformer,
  title={Bevformer: learning bird's-eye-view representation from lidar-camera via spatiotemporal transformers},
  author={Li, Zhiqi and Wang, Wenhai and Li, Hongyang and Xie, Enze and Sima, Chonghao and Lu, Tong and Yu, Qiao and Dai, Jifeng},
  journal={IEEE Transactions on Pattern Analysis and Machine Intelligence},
  year={2024},
  publisher={IEEE}
}

@article{shi2024mtr++,
  title={Mtr++: Multi-agent motion prediction with symmetric scene modeling and guided intention querying},
  author={Shi, Shaoshuai and Jiang, Li and Dai, Dengxin and Schiele, Bernt},
  journal={IEEE Transactions on Pattern Analysis and Machine Intelligence},
  volume={46},
  number={5},
  pages={3955--3971},
  year={2024},
  publisher={IEEE}
}

@article{zhou2023qcnext,
  title={Qcnext: A next-generation framework for joint multi-agent trajectory prediction},
  author={Zhou, Zikang and Wen, Zihao and Wang, Jianping and Li, Yung-Hui and Huang, Yu-Kai},
  journal={arXiv preprint arXiv:2306.10508},
  year={2023}
}

@article{huang2023differentiable,
  title={Differentiable integrated motion prediction and planning with learnable cost function for autonomous driving},
  author={Huang, Zhiyu and Liu, Haochen and Wu, Jingda and Lv, Chen},
  journal={IEEE transactions on neural networks and learning systems},
  year={2023},
  publisher={IEEE}
}

@inproceedings{huang2023gameformer,
  title={Gameformer: Game-theoretic modeling and learning of transformer-based interactive prediction and planning for autonomous driving},
  author={Huang, Zhiyu and Liu, Haochen and Lv, Chen},
  booktitle={Proceedings of the IEEE/CVF International Conference on Computer Vision},
  pages={3903--3913},
  year={2023}
}

@inproceedings{wang2022detr3d,
  title={Detr3d: 3d object detection from multi-view images via 3d-to-2d queries},
  author={Wang, Yue and Guizilini, Vitor Campagnolo and Zhang, Tianyuan and Wang, Yilun and Zhao, Hang and Solomon, Justin},
  booktitle={Conference on Robot Learning},
  pages={180--191},
  year={2022},
  organization={PMLR}
}

@misc{li2024womd,
      title={WOMD-Reasoning: A Large-Scale Dataset for Interaction Reasoning in Driving}, 
      author={Yiheng Li and Cunxin Fan and Chongjian Ge and Zhihao Zhao and Chenran Li and Chenfeng Xu and Huaxiu Yao and Masayoshi Tomizuka and Bolei Zhou and Chen Tang and Mingyu Ding and Wei Zhan},
      year={2025},
      eprint={2407.04281},
      archivePrefix={arXiv},
      primaryClass={cs.RO},
      url={https://arxiv.org/abs/2407.04281}, 
}

@article{zhou2024v2xpnp,
  title={{V2XPnP}: Vehicle-to-Everything Spatio-Temporal Fusion for Multi-Agent Perception and Prediction},
  author={Zhou, Zewei and Xiang, Hao and Zheng, Zhaoliang and Zhao, Seth Z and Lei, Mingyue and Zhang, Yun and Cai, Tianhui and Liu, Xinyi and Liu, Johnson and Bajji, Maheswari and others},
  journal={arXiv preprint arXiv:2412.01812},
  year={2024}
}

@article{liu2025hybrid,
  title={Hybrid-Prediction Integrated Planning for Autonomous Driving},
  author={Liu, Haochen and Huang, Zhiyu and Huang, Wenhui and Yang, Haohan and Mo, Xiaoyu and Lv, Chen},
  journal={IEEE Transactions on Pattern Analysis and Machine Intelligence},
  year={2025},
  publisher={IEEE}
}

@article{liang2022bevfusion,
  title={Bevfusion: A simple and robust lidar-camera fusion framework},
  author={Liang, Tingting and Xie, Hongwei and Yu, Kaicheng and Xia, Zhongyu and Lin, Zhiwei and Wang, Yongtao and Tang, Tao and Wang, Bing and Tang, Zhi},
  journal={Advances in Neural Information Processing Systems},
  volume={35},
  pages={10421--10434},
  year={2022}
}

@inproceedings{jia2023think,
  title={Think twice before driving: Towards scalable decoders for end-to-end autonomous driving},
  author={Jia, Xiaosong and Wu, Penghao and Chen, Li and Xie, Jiangwei and He, Conghui and Yan, Junchi and Li, Hongyang},
  booktitle={Proceedings of the IEEE/CVF Conference on Computer Vision and Pattern Recognition},
  pages={21983--21994},
  year={2023}
}

@inproceedings{shao2023reasonnet,
  title={Reasonnet: End-to-end driving with temporal and global reasoning},
  author={Shao, Hao and Wang, Letian and Chen, Ruobing and Waslander, Steven L and Li, Hongsheng and Liu, Yu},
  booktitle={Proceedings of the IEEE/CVF conference on computer vision and pattern recognition},
  pages={13723--13733},
  year={2023}
}

@inproceedings{pan2024vlp,
  title={Vlp: Vision language planning for autonomous driving},
  author={Pan, Chenbin and Yaman, Burhaneddin and Nesti, Tommaso and Mallik, Abhirup and Allievi, Alessandro G and Velipasalar, Senem and Ren, Liu},
  booktitle={Proceedings of the IEEE/CVF Conference on Computer Vision and Pattern Recognition},
  pages={14760--14769},
  year={2024}
}

@InProceedings{Wang_2024_CVPR,
    author    = {Wang, Yuqi and He, Jiawei and Fan, Lue and Li, Hongxin and Chen, Yuntao and Zhang, Zhaoxiang},
    title     = {Driving into the Future: Multiview Visual Forecasting and Planning with World Model for Autonomous Driving},
    booktitle = {Proceedings of the IEEE/CVF Conference on Computer Vision and Pattern Recognition (CVPR)},
    month     = {June},
    year      = {2024},
    pages     = {14749-14759}
}

@inproceedings{hu2022st,
  title={St-p3: End-to-end vision-based autonomous driving via spatial-temporal feature learning},
  author={Hu, Shengchao and Chen, Li and Wu, Penghao and Li, Hongyang and Yan, Junchi and Tao, Dacheng},
  booktitle={European Conference on Computer Vision},
  pages={533--549},
  year={2022},
  organization={Springer}
}

@article{zhou2024vision,
  title={Vision language models in autonomous driving: A survey and outlook},
  author={Zhou, Xingcheng and Liu, Mingyu and Yurtsever, Ekim and Zagar, Bare Luka and Zimmer, Walter and Cao, Hu and Knoll, Alois C},
  journal={IEEE Transactions on Intelligent Vehicles},
  year={2024},
  publisher={IEEE}
}

@inproceedings{fang2023tbp,
  title={Tbp-former: Learning temporal bird's-eye-view pyramid for joint perception and prediction in vision-centric autonomous driving},
  author={Fang, Shaoheng and Wang, Zi and Zhong, Yiqi and Ge, Junhao and Chen, Siheng},
  booktitle={Proceedings of the IEEE/CVF conference on computer vision and pattern recognition},
  pages={1368--1378},
  year={2023}
}

@inproceedings{jin2023adapt,
  title={Adapt: Action-aware driving caption transformer},
  author={Jin, Bu and Liu, Xinyu and Zheng, Yupeng and Li, Pengfei and Zhao, Hao and Zhang, Tong and Zheng, Yuhang and Zhou, Guyue and Liu, Jingjing},
  booktitle={2023 IEEE International Conference on Robotics and Automation (ICRA)},
  pages={7554--7561},
  year={2023},
  organization={IEEE}
}

@inproceedings{fu2024drive,
  title={Drive like a human: Rethinking autonomous driving with large language models},
  author={Fu, Daocheng and Li, Xin and Wen, Licheng and Dou, Min and Cai, Pinlong and Shi, Botian and Qiao, Yu},
  booktitle={2024 IEEE/CVF Winter Conference on Applications of Computer Vision Workshops (WACVW)},
  pages={910--919},
  year={2024},
  organization={IEEE}
}

@inproceedings{li2024ego,
  title={Is ego status all you need for open-loop end-to-end autonomous driving?},
  author={Li, Zhiqi and Yu, Zhiding and Lan, Shiyi and Li, Jiahan and Kautz, Jan and Lu, Tong and Alvarez, Jose M},
  booktitle={Proceedings of the IEEE/CVF Conference on Computer Vision and Pattern Recognition},
  pages={14864--14873},
  year={2024}
}

@article{ding2024hint,
  title={Hint-ad: Holistically aligned interpretability in end-to-end autonomous driving},
  author={Ding, Kairui and Chen, Boyuan and Su, Yuchen and Gao, Huan-ang and Jin, Bu and Sima, Chonghao and Zhang, Wuqiang and Li, Xiaohui and Barsch, Paul and Li, Hongyang and others},
  journal={arXiv preprint arXiv:2409.06702},
  year={2024}
}

@article{jaech2024openai,
  title={Openai o1 system card},
  author={Jaech, Aaron and Kalai, Adam and Lerer, Adam and Richardson, Adam and El-Kishky, Ahmed and Low, Aiden and Helyar, Alec and Madry, Aleksander and Beutel, Alex and Carney, Alex and others},
  journal={arXiv preprint arXiv:2412.16720},
  year={2024}
}

@article{ouyang2022training,
  title={Training language models to follow instructions with human feedback},
  author={Ouyang, Long and Wu, Jeffrey and Jiang, Xu and Almeida, Diogo and Wainwright, Carroll and Mishkin, Pamela and Zhang, Chong and Agarwal, Sandhini and Slama, Katarina and Ray, Alex and others},
  journal={Advances in neural information processing systems},
  volume={35},
  pages={27730--27744},
  year={2022}
}

@article{rafailov2023direct,
  title={Direct preference optimization: Your language model is secretly a reward model},
  author={Rafailov, Rafael and Sharma, Archit and Mitchell, Eric and Manning, Christopher D and Ermon, Stefano and Finn, Chelsea},
  journal={Advances in Neural Information Processing Systems},
  volume={36},
  pages={53728--53741},
  year={2023}
}

@article{guo2025deepseek,
  title={Deepseek-r1: Incentivizing reasoning capability in llms via reinforcement learning},
  author={Guo, Daya and Yang, Dejian and Zhang, Haowei and Song, Junxiao and Zhang, Ruoyu and Xu, Runxin and Zhu, Qihao and Ma, Shirong and Wang, Peiyi and Bi, Xiao and others},
  journal={arXiv preprint arXiv:2501.12948},
  year={2025}
}

@article{wang2023drivemlm,
  title={Drivemlm: Aligning multi-modal large language models with behavioral planning states for autonomous driving},
  author={Wang, Wenhai and Xie, Jiangwei and Hu, ChuanYang and Zou, Haoming and Fan, Jianan and Tong, Wenwen and Wen, Yang and Wu, Silei and Deng, Hanming and Li, Zhiqi and others},
  journal={arXiv preprint arXiv:2312.09245},
  year={2023}
}

@article{liao2025cot,
  title={Cot-drive: Efficient motion forecasting for autonomous driving with llms and chain-of-thought prompting},
  author={Liao, Haicheng and Kong, Hanlin and Wang, Bonan and Wang, Chengyue and Ye, Wang and He, Zhengbing and Xu, Chengzhong and Li, Zhenning},
  journal={arXiv preprint arXiv:2503.07234},
  year={2025}
}

@article{hegde2025distilling,
  title={Distilling Multi-modal Large Language Models for Autonomous Driving},
  author={Hegde, Deepti and Yasarla, Rajeev and Cai, Hong and Han, Shizhong and Bhattacharyya, Apratim and Mahajan, Shweta and Liu, Litian and Garrepalli, Risheek and Patel, Vishal M and Porikli, Fatih},
  journal={arXiv preprint arXiv:2501.09757},
  year={2025}
}

@article{zheng2024gaussianad,
  title={GaussianAD: Gaussian-Centric End-to-End Autonomous Driving},
  author={Zheng, Wenzhao and Wu, Junjie and Zheng, Yao and Zuo, Sicheng and Xie, Zixun and Yang, Longchao and Pan, Yong and Hao, Zhihui and Jia, Peng and Lang, Xianpeng and others},
  journal={arXiv preprint arXiv:2412.10371},
  year={2024}
}

@article{liu2025vlm,
  title={VLM-E2E: Enhancing End-to-End Autonomous Driving with Multimodal Driver Attention Fusion},
  author={Liu, Pei and Liu, Haipeng and Liu, Haichao and Liu, Xin and Ni, Jinxin and Ma, Jun},
  journal={arXiv preprint arXiv:2502.18042},
  year={2025}
}

@article{tian2024tokenize,
  title={Tokenize the world into object-level knowledge to address long-tail events in autonomous driving},
  author={Tian, Ran and Li, Boyi and Weng, Xinshuo and Chen, Yuxiao and Schmerling, Edward and Wang, Yue and Ivanovic, Boris and Pavone, Marco},
  journal={arXiv preprint arXiv:2407.00959},
  year={2024}
}

@misc{waywe2024lingo,
  title={LINGO-2: Driving with Natural Language},
  author={Waywe Research Team and others},
  year={2024}
}

@misc{hung2025norasmallopensourcedgeneralist,
      title={NORA: A Small Open-Sourced Generalist Vision Language Action Model for Embodied Tasks}, 
      author={Chia-Yu Hung and Qi Sun and Pengfei Hong and Amir Zadeh and Chuan Li and U-Xuan Tan and Navonil Majumder and Soujanya Poria},
      year={2025},
      eprint={2504.19854},
      archivePrefix={arXiv},
      primaryClass={cs.RO},
      url={https://arxiv.org/abs/2504.19854}, 
}

@article{ma2025reasoning,
  title={Reasoning Models Can Be Effective Without Thinking},
  author={Ma, Wenjie and He, Jingxuan and Snell, Charlie and Griggs, Tyler and Min, Sewon and Zaharia, Matei},
  journal={arXiv preprint arXiv:2504.09858},
  year={2025}
}

@misc{qiao2025lightemmalightweightendtoendmultimodal,
      title={LightEMMA: Lightweight End-to-End Multimodal Model for Autonomous Driving}, 
      author={Zhijie Qiao and Haowei Li and Zhong Cao and Henry X. Liu},
      year={2025},
      eprint={2505.00284},
      archivePrefix={arXiv},
      primaryClass={cs.RO},
      url={https://arxiv.org/abs/2505.00284}, 
}

@inproceedings{wu2025language,
  title={Language prompt for autonomous driving},
  author={Wu, Dongming and Han, Wencheng and Liu, Yingfei and Wang, Tiancai and Xu, Cheng-zhong and Zhang, Xiangyu and Shen, Jianbing},
  booktitle={Proceedings of the AAAI Conference on Artificial Intelligence},
  volume={39},
  number={8},
  pages={8359--8367},
  year={2025}
}

@inproceedings{qian2024nuscenes,
  title={Nuscenes-qa: A multi-modal visual question answering benchmark for autonomous driving scenario},
  author={Qian, Tianwen and Chen, Jingjing and Zhuo, Linhai and Jiao, Yang and Jiang, Yu-Gang},
  booktitle={Proceedings of the AAAI Conference on Artificial Intelligence},
  volume={38},
  number={5},
  pages={4542--4550},
  year={2024}
}

@inproceedings{nie2024reason2drive,
  title={Reason2drive: Towards interpretable and chain-based reasoning for autonomous driving},
  author={Nie, Ming and Peng, Renyuan and Wang, Chunwei and Cai, Xinyue and Han, Jianhua and Xu, Hang and Zhang, Li},
  booktitle={European Conference on Computer Vision},
  pages={292--308},
  year={2024},
  organization={Springer}
}

@inproceedings{ma2024dolphins,
  title={Dolphins: Multimodal language model for driving},
  author={Ma, Yingzi and Cao, Yulong and Sun, Jiachen and Pavone, Marco and Xiao, Chaowei},
  booktitle={European Conference on Computer Vision},
  pages={403--420},
  year={2024},
  organization={Springer}
}

@article{winter2025bevdriver,
  title={BEVDriver: Leveraging BEV Maps in LLMs for Robust Closed-Loop Driving},
  author={Winter, Katharina and Azer, Mark and Flohr, Fabian B},
  journal={arXiv preprint arXiv:2503.03074},
  year={2025}
}

@article{hu2022lora,
  title={Lora: Low-rank adaptation of large language models.},
  author={Hu, Edward J and Shen, Yelong and Wallis, Phillip and Allen-Zhu, Zeyuan and Li, Yuanzhi and Wang, Shean and Wang, Lu and Chen, Weizhu and others},
  journal={ICLR},
  volume={1},
  number={2},
  pages={3},
  year={2022}
}

@misc{liu2025dsdrivedistillinglargelanguage,
      title={DSDrive: Distilling Large Language Model for Lightweight End-to-End Autonomous Driving with Unified Reasoning and Planning}, 
      author={Wenru Liu and Pei Liu and Jun Ma},
      year={2025},
      eprint={2505.05360},
      archivePrefix={arXiv},
      primaryClass={cs.RO},
      url={https://arxiv.org/abs/2505.05360}, 
}

@article{tian2025nuscenes,
  title={NUSCENES-SPATIALQA: A Spatial Understanding and Reasoning Benchmark for Vision-Language Models in Autonomous Driving},
  author={Tian, Kexin and Mao, Jingrui and Zhang, Yunlong and Jiang, Jiwan and Zhou, Yang and Tu, Zhengzhong},
  journal={arXiv preprint arXiv:2504.03164},
  year={2025}
}

@misc{wang2025generativeaiautonomousdriving,
      title={Generative AI for Autonomous Driving: Frontiers and Opportunities}, 
      author={Yuping Wang and Shuo Xing and Cui Can and Renjie Li and Hongyuan Hua and Kexin Tian and Zhaobin Mo and Xiangbo Gao and Keshu Wu and Sulong Zhou and Hengxu You and Juntong Peng and Junge Zhang and Zehao Wang and Rui Song and Mingxuan Yan and Walter Zimmer and Xingcheng Zhou and Peiran Li and Zhaohan Lu and Chia-Ju Chen and Yue Huang and Ryan A. Rossi and Lichao Sun and Hongkai Yu and Zhiwen Fan and Frank Hao Yang and Yuhao Kang and Ross Greer and Chenxi Liu and Eun Hak Lee and Xuan Di and Xinyue Ye and Liu Ren and Alois Knoll and Xiaopeng Li and Shuiwang Ji and Masayoshi Tomizuka and Marco Pavone and Tianbao Yang and Jing Du and Ming-Hsuan Yang and Hua Wei and Ziran Wang and Yang Zhou and Jiachen Li and Zhengzhong Tu},
      year={2025},
      eprint={2505.08854},
      archivePrefix={arXiv},
      primaryClass={cs.CV},
      url={https://arxiv.org/abs/2505.08854}, 
}

@article{brohan2022rt,
  title={Rt-1: Robotics transformer for real-world control at scale},
  author={Brohan, Anthony and Brown, Noah and Carbajal, Justice and Chebotar, Yevgen and Dabis, Joseph and Finn, Chelsea and Gopalakrishnan, Keerthana and Hausman, Karol and Herzog, Alex and Hsu, Jasmine and others},
  journal={arXiv preprint arXiv:2212.06817},
  year={2022}
}

@article{xu2025wod,
  title={WOD-E2E: Waymo Open Dataset for End-to-End Driving in Challenging Long-tail Scenarios},
  author={Xu, Runsheng and Lin, Hubert and Jeon, Wonseok and Feng, Hao and Zou, Yuliang and Sun, Liting and Gorman, John and Tolstaya, Kate and Tang, Sarah and White, Brandyn and Ben, Sapp and Tan, Mingxing and Hwang, Jyh-Jing and Anguelov, Dragomir},
  journal={arXiv preprint arXiv:2510.26125},
  year={2025}
}
